\documentclass{article}

\PassOptionsToPackage{numbers, sort}{natbib}
\usepackage[preprint]{neurips_2023}

\usepackage[utf8]{inputenc} %
\usepackage[T1]{fontenc}    %
\usepackage{hyperref}       %
\usepackage{url}            %
\usepackage{booktabs}       %
\usepackage{amsfonts}       %
\usepackage{nicefrac}       %
\usepackage{microtype}      %
\usepackage{xcolor}         %
\usepackage{graphicx}
\usepackage{wrapfig}
\usepackage{booktabs}
\usepackage{multirow}
\usepackage{caption}
\usepackage{subcaption}
\usepackage{enumitem}
\usepackage{float}
\usepackage{algorithm}
\usepackage{algpseudocode}

\usepackage{amsmath}
\usepackage{amssymb}
\usepackage{mathtools}
\usepackage{amsthm}

\theoremstyle{plain}
\newtheorem{theorem}{Theorem}[section]
\newtheorem{proposition}[theorem]{Proposition}

\newtheorem{definition}[theorem]{Definition}

\theoremstyle{remark}

\title{Diffusion Denoising Process for Perceptron Bias in Out-of-distribution Detection}

\author{%
  Luping Liu$^1$, Yi Ren$^1$, Xize Cheng$^1$, Rongjie Huang$^1$, Chongxuan Li$^2$, Zhou Zhao$^1$\thanks{Corresponding author}\\
  $^1$Zhejiang University\ \  $^2$Renmin University of China\\
  \texttt{\{luping.liu, rayeren, chengxize, rongjiehuang, zhaozhou\}@zju.edu.cn,} \\\texttt{chongxuanli@ruc.edu.cn}
   \\
}

\begin{document}

\maketitle

\begin{abstract}
  Out-of-distribution (OOD) detection is a crucial task for ensuring the reliability and safety of deep learning. Currently, discriminator models outperform other methods in this regard. However, the feature extraction process used by discriminator models suffers from the loss of critical information, leaving room for bad cases and malicious attacks. In this paper, we introduce a new perceptron bias assumption that suggests discriminator models are more sensitive to certain features of the input, leading to the overconfidence problem. To address this issue, we propose a novel framework that combines discriminator and generation models and integrates diffusion models (DMs) into OOD detection. We demonstrate that the diffusion denoising process (DDP) of DMs serves as a novel form of asymmetric interpolation, which is well-suited to enhance the input and mitigate the overconfidence problem. The discriminator model features of OOD data exhibit sharp changes under DDP, and we utilize the norm of this change as the indicator score. Our experiments on CIFAR10, CIFAR100, and ImageNet show that our method outperforms SOTA approaches. Notably, for the challenging InD ImageNet and OOD species datasets, our method achieves an AUROC of 85.7, surpassing the previous SOTA method's score of 77.4. Our implementation is available at \url{https://github.com/luping-liu/DiffOOD}.
\end{abstract}

\section{Introduction}
\label{sec_intro}

\begin{wrapfigure}{r}{0.3\linewidth}
  \centering
  \vspace*{-0.3cm}
  \includegraphics[width=\linewidth]{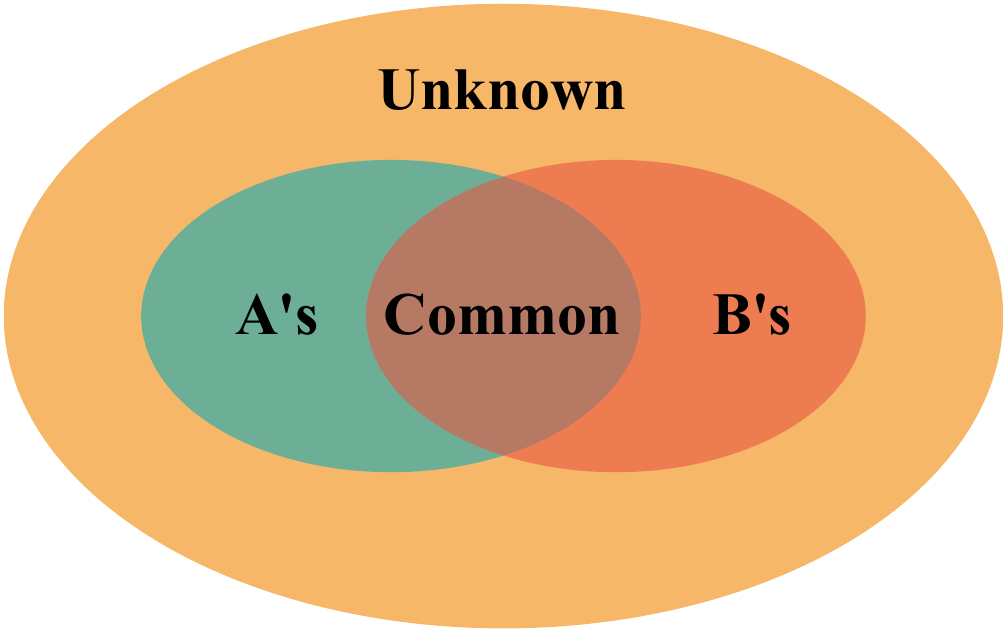}
  \includegraphics[width=\linewidth, trim=0 15 0 -10]{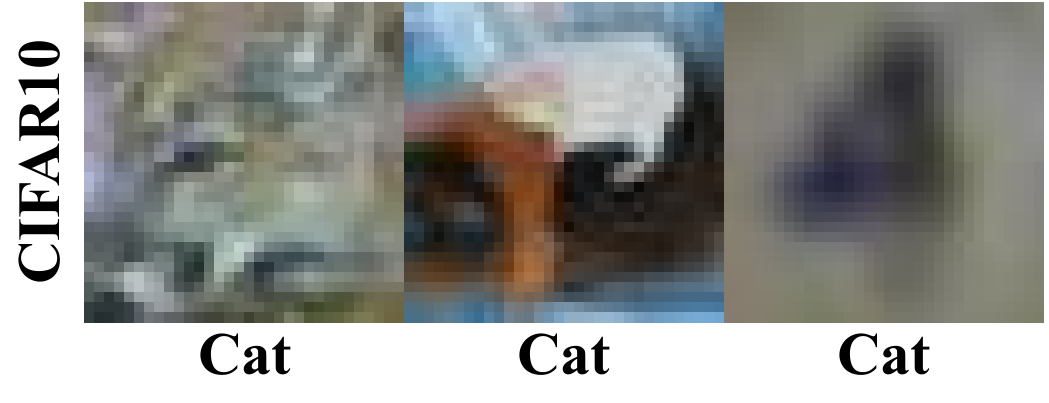}
  \caption{An overview of the feature space and overconfidence examples of discriminator models.}
  \label{fig_over}
  \vspace*{-0.7cm}
\end{wrapfigure}

Out-of-distribution (OOD) detection is a crucial task for deep learning models, as it helps them determine their capability boundary and prevents them from being deceived by OOD data. This task is closely related to many real-world machine learning applications, such as cybersecurity \citep{xin2018machine}, medical diagnosis \citep{latif2018phonocardiographic, guo2020ensemble}, and autopilot \citep{geiger2012we}. Existing methods for OOD detection can be broadly categorized into discriminator-based and generation-based approaches. Discriminator-based methods \citep{wang2022vim} utilize either the logit or feature space to perform detection, while generation-based methods \citep{an2015variational, nalisnick2019detecting} rely on reconstruction differences in data space or density estimation in latent space. 

Discriminator-based methods are generally faster and more effective in extracting useful features for OOD detection. However, \citet{nguyen2015deep} show that some OOD data can be misclassified with high confidence, resulting in the overconfidence problem. On the other hand, generation-based methods can capture the entire distribution 

but often lack effective indicator scores to compete with state-of-the-art discriminator-based methods.  %

\begin{figure}
    \centering
    \includegraphics[width=\linewidth, trim=5 5 5 0]{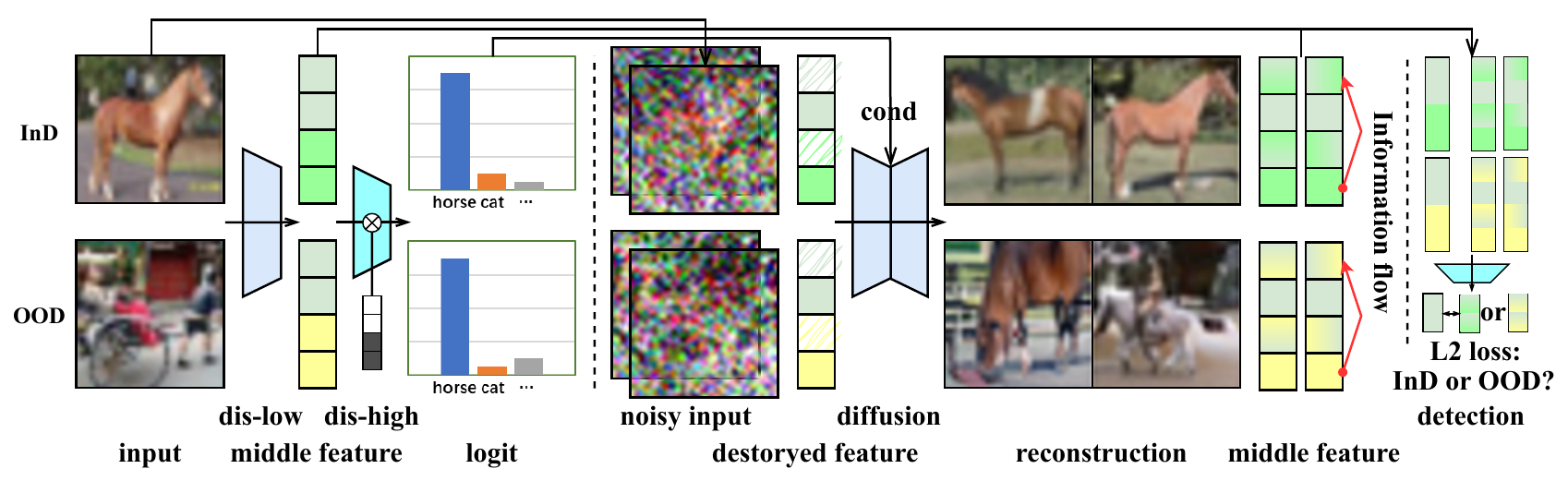}
    \caption{The pipeline of our OOD detection method. The discriminator model can identify more information through the flow of information during the diffusion denoising process. Here, We use a mask to represent the information loss in the discriminator model.} %
    \label{fig_pipeline}
    \vspace*{-0.4cm}
\end{figure}

In this paper, we propose a novel approach that combines discriminator and generation models to counteract each other's limitations. Our analysis reveals that discriminator models do not assign significant importance to common features of different categories, which may be important for OOD detection. The relationship between common and exclusive features is depicted in Figure \ref{fig_over}. Thus, we introduce a new assumption that discriminator models focus on specific features of the input, leading to the overconfidence problem. For example, the model focuses on certain patterns in Figure \ref{fig_over} and \ref{fig_pipeline}, both misclassifying OOD data with high confidence. Conversely, generation models can capture the entire data distribution, including the common features of different categories. Therefore, we use them to enhance the input and help the discriminator models. This approach is analogous to the principle behind smoke alarms. Because the air is flowing, we can install a fixed smoke alarm to monitor the whole house, while generation models make information ``flow'' more effectively.

To develop effective generation strategies, we incorporate diffusion models (DMs) into OOD detection. DMs have achieved state-of-the-art results in many generation tasks, including \citet{vahdat2021score} and \citet{ho2022video}. We observe that the diffusion denoising process (DDP) of DMs serves as a novel form of asymmetric interpolation, functioning as both a reconstruction and interpolation process. The reconstruction property enables DDP to enhance the input and enable it to ``flow', while the interpolation property helps to maintain control over the adjustments. Our method employs a two-step framework for OOD detection. First, we utilize DDP to generate additional samples based on the input. Next, we input all of these samples into a pre-trained discriminator model. Anomalies previously ignored by the discriminator model are exposed in the flow of information formed by DDP, and then the feature results of the discriminator model change sharply. Therefore, we utilize the norm of the dynamic change in features as the indicator score. The pipeline is illustrated in Figure \ref{fig_pipeline}.

We choose six representative methods to compare with our methods on CIFAR10, CIFAR100, and ImageNet. According to our experiments, our methods outperform existing methods. Notably, for the challenging InD ImageNet and OOD species datasets, our method achieves an AUROC of 85.7, surpassing the previous SOTA method's score of 77.4. Our work has the following contributions:

\begin{itemize}[leftmargin=*]
  \item We propose a new perceptron bias assumption that explains the overconfidence problem, suggesting that discriminator models are more sensitive to certain features of the input. 
  \item  We introduce diffusion models into OOD detection and demonstrate that the diffusion denoising process of invertible DMs serves as a novel form of asymmetric interpolation, enabling effective enhancement of input data while maintaining control in the denoising process.
  \item  We design a two-step detection method that utilizes a ResNet to extract features and the diffusion denoising process to mitigate overconfidence issues. Our method is highly explicable and achieves state-of-the-art results.
\end{itemize}

\section{Background}
\label{sec_back}

In this section, we first introduce existing methods for OOD detection. Then, we show the development of diffusion models related to our paper. Due to the limited space in the main paper, more related works about diffusion models and similarities and differences between our method and existing methods can be found in Appendix \ref{app_related}.

\subsection{Out-of-distribution Detection}
\label{sec_ood_bg}

OOD detection is an important task that can help neural networks to determine their capability boundary. More specifically, let $X = \{x_1, \dots, x_n\} \sim p$ be a group of images from the in-distribution (InD) $p$. We want to build a detector $f$ that $f(x_1,\dots,x_n)=1, \forall i, p(x_i)\geq \sigma$ and $f(x_1,\dots,x_n)=0, \forall i, p(x_i)\leq \sigma$. Here, $\sigma$ controls the decision boundary. When we get another group of data $Y = \{y_1, \dots, y_n\}$, we decide whether this group is from InD $p$ or an unknown distribution $q$ based on the results of $f$. If $n=1$, this is pointwise OOD detection, and if $n\geq 2$, this is group OOD detection. In general, the existing OOD detection methods can be categorized into discriminator-based and generation-based methods. %

Discriminator-based methods design indicator scores based on the output of discriminator models. Some methods can be used without modifying the model. ODIN \citep{liang2018enhancing} uses temperature scaling and the softmax results to detect OOD samples. ViM \citep{wang2022vim} combines the information of features and logits. KNN \citep{sun2022out} includes the kth nearest neighbor of the input in feature space into the detection process. Some methods try to improve the detection ability in the training process. G-ODIN \citep{hsu2020generalized} designs a new loss function. ConfGAN \citep{sricharan2018building} generates OOD data using GANs to help the discriminator models to determine the boundary. PixMix \citep{hendrycks2022pixmix} uses data augmentation to improve the detection results. SSD \citep{sehwag2020ssd} uses self-supervised learning to improve feature extraction. 

Generation-based methods employ the difference in reconstruction between the input space or the density estimation in the latent space for OOD detection. The former assumes that the generation models can reconstruct the in-distribution data more effectively, while the latter utilizes the distribution transformation capability of generation models to convert the input distribution into a simple Gaussian distribution. In more detail, the likelihood of the input is a direct indicator score, but \citet{nalisnick2018deep} find that OOD data can also locate in the high-likelihood area. \citet{nalisnick2019detecting} find the InD data is concentrated in the typical set instead of the high likelihood area and design new methods using the typical set. \citet{serra2019input} find that we can use input complexity to correct the bias of likelihood. In addition to the likelihood, many existing statistical methods can detect whether a distribution obeys standard Gaussian distribution. \citet{zhang2020out} uses KL-divergence to detect OOD data. \citet{jiang2022revisiting} use a nonparametric statistics method, the Kolmogorov-Smirnov test.

\subsection{Diffusion Model}

\paragraph{Classical diffusion model}

DMs build a transformation from Gaussian distribution to image distribution through a multistep denoising process. Given a data distribution $x_0 \sim q(x_0)$, the diffusion process satisfies a Markov process as following \cite{ho2020denoising}:
\begin{equation}
  \begin{split}
     &q(x_{1:T}|x_0) = \prod_{t=1}^T \mathcal{N}(\sqrt{1 - \beta_t} x_{t-1}, \beta_t I) \\
     &q(x_t| x_{0}) = \mathcal{N}(\sqrt{\bar{\alpha}_t}x_0, (1-\bar{\alpha}_t) I).
  \end{split}
  \label{eq_diffusion}
\end{equation}
Here, $T=1000$, which is the max iteration step. $\beta_t\in (0, 1)$, which controls the speed of adding noise. Additionally, $\alpha_t=1-\beta_t$, $\bar{\alpha}_t = \prod_{i=1}^t\alpha_i$, $\bar{\mu}_t = \frac{\sqrt{\bar{\alpha}_{t-1}}\beta_t}{1-\bar{\alpha}_t}x_0 + \frac{\sqrt{\alpha_t}(1-\bar{\alpha}_{t-1})}{1-\bar{\alpha}_t}x_t$ and $\bar{\beta}_t=\frac{1-\bar{\alpha}_{t-1}}{1-\bar{\alpha}_t}\beta_t$. The objective function is defined by: 
\begin{equation}
    \small
   \begin{split}
      L_{t}  = \mathbb{E}_{x_0, \epsilon}\left[\frac{\beta_t^2}{\alpha_t(1-\bar{\alpha}_t)}||\epsilon - \epsilon_\theta(\sqrt{\bar{\alpha}_t}x_0+ \sqrt{1-\bar{\alpha}_t}\epsilon, t)||^2\right].
   \end{split}
   \label{eq_ob_func}
\end{equation}
Here, $\epsilon_\theta$ is an estimate of the noise $\epsilon$. After we get well-trained $\epsilon_\theta$, according to \citet{song2021denoising}, the denoising process of Denoising Diffusion Probabilistic Models (DDPMs) and Denoising Diffusion Implicit Models (DDIMs) satisfies:
\begin{equation}
    x_{t-\delta} = \sqrt{\bar{\alpha}_{t-\delta}}\left(\frac{x_t-\sqrt{1-\bar{\alpha}_t}\epsilon_\theta(x_t, t)}{\sqrt{\bar{\alpha}_t}}\right) + \sqrt{1-\bar{\alpha}_{t-\delta}-\sigma^2_t}\epsilon_\theta(x_t, t) + \sigma_t \epsilon_t.
   \label{eq_ddpm}
\end{equation}

Here, $\delta$ is the iteration step size. If $\sigma_t$ equals one, Equation (\ref{eq_ddpm}) represents the denoising process of DDPMs; if $\sigma_t$ equals zero, this equation represents the denoising process of DDIMs. In Appendix \ref{app_inv_diff}, we will further describe how to make the iteration of the diffusion model fast and invertible.

\textbf{Classifier-free guidance} \citet{ho2021classifier} show a simple and effective way to generate conditional samples called classifier-free guidance. It adds a condition embedding $c$ into the model during the training process and changes the estimation of noise as:
\begin{equation}
  \bar{\epsilon}_\theta(x_t, c, t) = \epsilon_\theta(x_t, \emptyset, t) + \omega \left(\epsilon_\theta(x_t, c, t) - \epsilon_\theta(x_t, \emptyset, t)\right).
\end{equation}
Here, $\omega$ is the guidance weight, which controls the balance between realness and diversity.

\section{Perceptron Bias and Diffusion Denoising Process}
\label{sec_assumption_process}

In this section, we propose a new perceptron bias assumption to explain the overconfidence of discriminator models and highlight the potential of generation models to address this issue. We then use a toy example to demonstrate how generation models can be used to reduce information loss and incorporate the diffusion denoising process (DDP) into OOD detection. We show that DDP serves as both a reconstruction and interpolation operator, making it well-suited to address the perceptron bias problem. We then visualize the impact of DDP on real-world scenarios and explain how to select indicator scores. Furthermore, we introduce conditional diffusion models to address multiclass detection and provide our algorithms.  %

\subsection{Why Generation Models are Essential?}

In the introduction, we briefly introduce the perceptron bias problem in discriminator models. Here, we provide a more detailed analysis of this issue. We find discriminator models are more sensitive to features that belong exclusively to certain categories and do not place as much value on common features. This is a reasonable decision for an ``smart'' discriminator model due to both the model structure and the training process.

In terms of the model structure, they employ convolutional and pooling layers to compress the data space. Consequently, the models must learn to eliminate some relatively useless information to achieve high classification accuracy. In terms of the training process, some features may not be crucial for the original classification problem. For instance, the eyes may not be a significant feature for distinguishing between dogs and cats. Consequently, the models may ignore this information. However, this feature is crucial in determining whether the input is animal or OOD data and should not be disregarded. In summary, the perceptron bias problem can be described as follows:

\begin{definition}
  The perceptron bias problem is that a fixed discriminator model focuses only on certain features of the input.
\end{definition}

These analyses highlight that a discriminator model is not omniscient and suffers from excessive information loss, particularly common features that belong to different categories. In contrast, generation models can effectively model information, including common features. This underscores the potential of generation models to enhance OOD detection.

\subsection{Toy Example}

To better illustrate the potential of generation models in addressing this issue, we simplify the perceptron bias to a constraint operator $f|_S$,\footnote{$S$ is the support area of $f$.} with the core idea of binarizing such bias into pass or fail. The data passes through a discriminator model as if it had been filtered through a mask, resulting in the loss of some information.

\begin{figure*}[t]
  \centering
  \begin{minipage}[t]{0.74\linewidth}
      \includegraphics[width=\linewidth, trim=210 170 210 135]{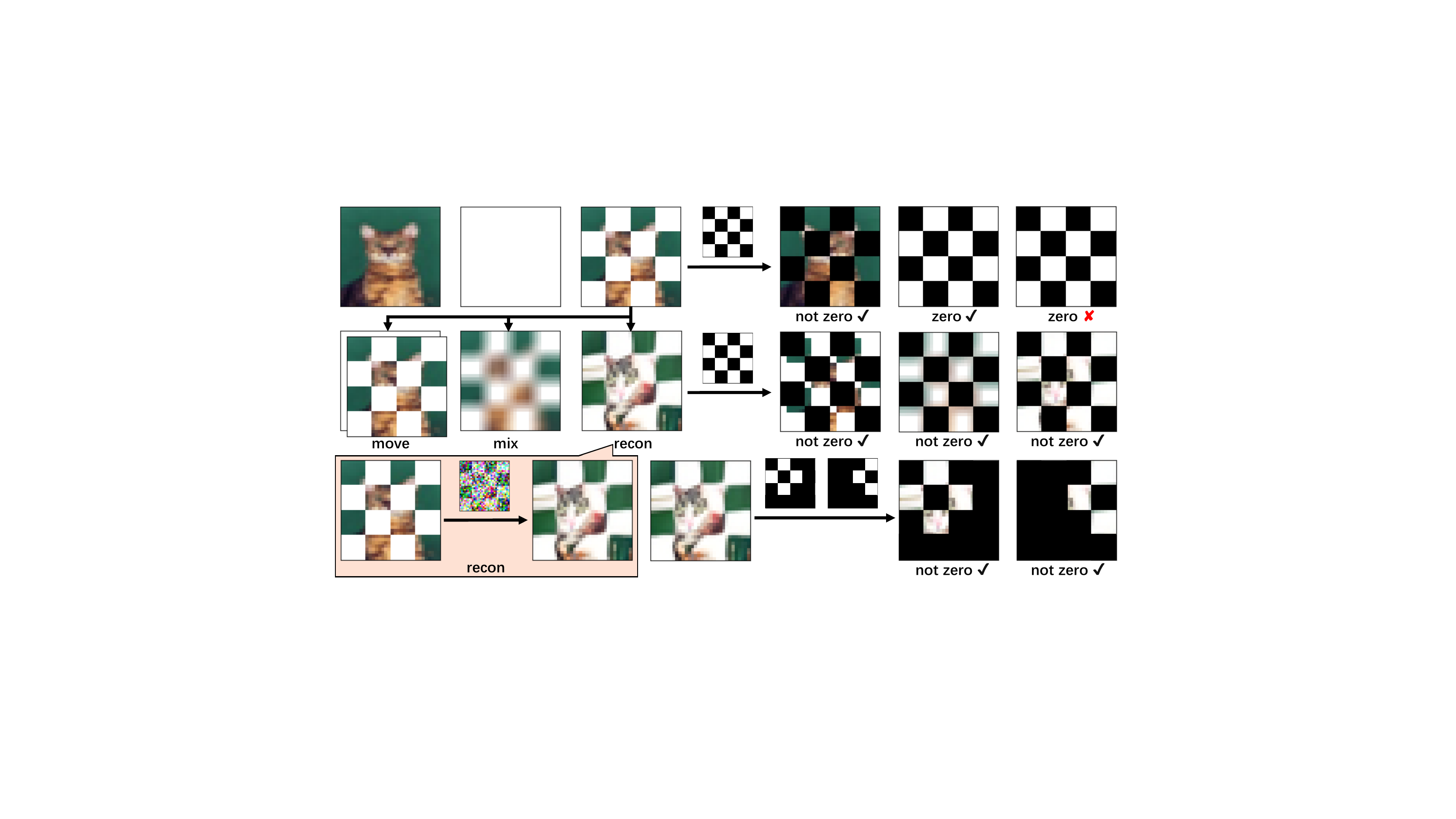}
      \vspace*{-0.2cm}
      \caption{A toy OOD detection process to detect whether the input is a neighbor of zero (the empty image). The top row shows two normal cases and a bad case. The middle row provides three different additional operators to correct this bad case. The bottom row shows that the above operators are suitable for different shapes of support areas.}
      \label{fig_toy}
  \end{minipage}
  \ \ 
  \begin{minipage}[t]{0.24\linewidth}
    \includegraphics[width=\linewidth, trim=5 8 0 0]{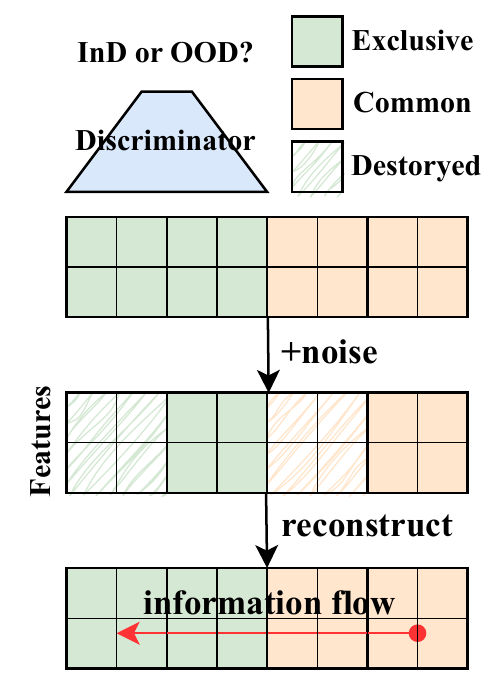}
    \vspace*{-0.2cm}
    \caption{Common features flow towards the exclusive area and then can be detected by the discriminator model.}
    \label{fig_recon}
  \end{minipage}
  \vspace*{-0.4cm}
\end{figure*}

Here, the input space is a 2-dimensional square space. The support area $S$ of the restriction operator $f$ is a collection of small patches $\bigcup_{i\in I_S} S_i$ at the pixel level. Our InD data is a uniform distribution in a spherical neighborhood of zero (the empty image), and our objective is to detect whether a new input is within the neighborhood of zero. We illustrate this toy example in Figure \ref{fig_toy}, where the top row shows two normal cases and one bad case under this setting. The bad case appears to be zero after the restriction operator $f$. In the middle row of Figure \ref{fig_toy}, we correct the result by using three additional operators: moving, mixing, and reconstruction.

The first two methods can perfectly solve the information loss problem in Figure \ref{fig_toy}, and the proof is provided in Appendix \ref{app_proof}. The key idea behind these methods is to move or mix information from one place to another. This allows the detector $f$ to obtain the complete information from a small support area $S$, thereby reducing the information loss. Moreover, these methods can be extended to support areas $S$ of any shape, as demonstrated in the right side of the bottom row of Figure \ref{fig_toy}. Therefore, we can design general operators for different $S$, which is essential as we cannot always obtain the exact $S$ in real-world scenarios.

Although the above two methods exemplify our primary concept, their capacity is restricted to the pixel level. To design a more general method, we propose a third reconstruction operator that first destroys images and then employs a generation model to reconstruct them, which is also a form of data augmentation. While we refer to it as reconstruction, it does not restore the input entirely but rather reconstructs the missing parts based on available input information. This creates a form of information ``flow'' from one place to another, such as moving (e.g., the cat's face moving to the left and down), mixing (e.g., the boundary of each small box becoming unclear), and semantic-level moving (e.g., the cat's color becoming lighter), as shown in Figure \ref{fig_toy}.

\textbf{Core Concept} In essence, the interplay between discriminator and generation models can be likened to that of smoke alarms and the flow of air. While a smoke alarm is stationary within a building, it is capable of monitoring the entire space due to the movement of air. Similarly, although the discriminator may focus on specific features, the generation models can facilitate the flow of information to enable effective OOD detection. We illustrate this process in Figure \ref{fig_recon}.

\subsection{Why Choose Diffusion Models?}

While all generation models have the potential to reconstruct damaged images, this strategy also has its weaknesses. As shown in Figure \ref{fig_toy}, we can detect OOD examples more accurately using a reconstruction operator. However, we also need to ensure that pure white pictures remain clean. This challenge highlights the need for more control over the reconstruction strategy. We find that diffusion models are particularly suitable for two main reasons.

\textbf{Controlled noise intensity} The first advantage of diffusion models is their ability to naturally control the intensity of the added noise, which in turn affects the reconstruction results. This is important because combining different intensities of noise would make the reconstruction much more complicated and challenging to design indicators for OOD detection.

\textbf{Controlled flow vector} The flow vector is the difference between the original input and the reconstruction output. We find that the diffusion denoising process (DDP) of diffusion models is a form of interpolation under invertible conditions. As such, it serves as both a reconstruction and interpolation operator. The fact that DDP is a reconstruction operator implies that it can preserve InD data relatively unchanged while moving OOD data towards the high-density area of InD, which demonstrates the control of the norm of the flow vector. The fact that DDP is an interpolation operator suggests that we can control the direction of change and maintain InD data changes within the InD area when timestep $t$ is large, which demonstrates the control of the direction of the flow vector. By combining norm and direction control, we can control the flow vector.

\begin{figure}[t]
    \begin{minipage}{.58\linewidth}
      \centering
      \includegraphics[width=\linewidth, trim=72 15 82 5]{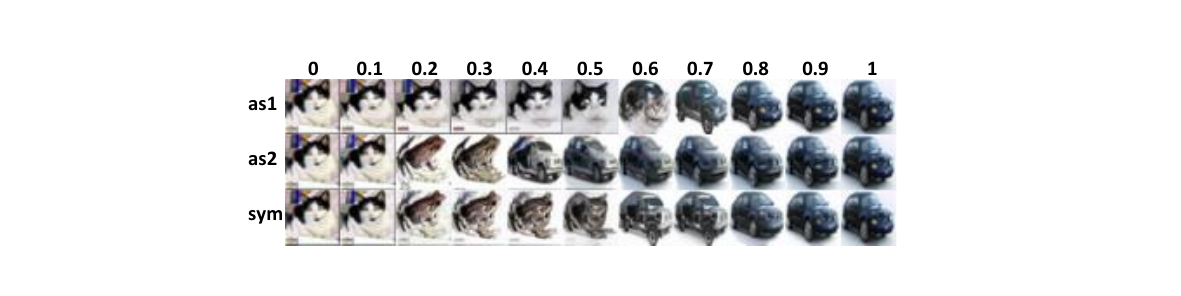}
      \caption{Interpolation results using two fixed images.}
      \label{fig_itp}
    \end{minipage}
    \ \ 
    \begin{minipage}{.4\linewidth}
      \begin{algorithm}[H]
        \caption{Asymmetric interpolation}
        \label{alg_itp2}
        \begin{algorithmic}[1]
            \State {\bfseries input:} Images $x_1, x_2$, interpolation timestep $t$
            \State $\epsilon_2 = \Phi(x_2, 0, T)$
            \State $x_t^{\text{inter}} = \sqrt{\bar{\alpha}_t} x_1 + \sqrt{1 - \bar{\alpha}_t} \epsilon_2 $
            \State $x^{\text{inter}} = \Phi(x_t^{\text{inter}}, t, 0)$
            \State {\bfseries return} $x^{\text{inter}}$
        \end{algorithmic}
    \end{algorithm}
    \end{minipage}
    \vspace*{-0.4cm}
\end{figure}

Many previous papers use diffusion models to interpolate two inputs, but our method is different from the existing symmetric approach using spherical linear interpolation \citep{shoemake1985animating}. We combine invertible diffusion models and DDP $\Phi$ to obtain a new asymmetric interpolation. The algorithm is shown in Algorithm \ref{alg_itp2}. Here, $\Phi(x, t_1, t_2)$ refers to the DDP process from $t_1$ to $t_2$. Setting $x_1$, $x_2$ are two images and $\epsilon_2=\Phi(x_2, 0, T)$ is the reverse noise of an image $x_2$, we use $x_1$ and $\epsilon_2$ to obtain $x_t^{\text{inter}}$, then denoise it to obtain the interpolation result $x^{\text{inter}}$. When $t$ equals zeros, we do not add any noise, and $x^{\text{inter}}_0 = x_1$, so the interpolation result is $x_1$. When $t$ equals $T$, we remove the total image $x_1$ and only leave $x^{\text{inter}}_T = \epsilon_2$, we will perform a full denoising process to this noise $\epsilon_2$. Because the denoising process is invertible, we can get the image $\Phi(\epsilon_2, T, 0)=x_2$. Therefore, the outputs of DDP gradually change from $x_1$ to $x_2$. Figure \ref{fig_itp} shows the three different interpolation results. The first row uses the cat as $x_1$, and the second row uses the car as $x_1$. We can find that $x_1$ can be better preserved compared to the symmetric interpolation in the third row.\footnote{More analysis and visual results about the new interpolation using DDP are in Appendix \ref{app_itp}.} In summary, we have:

\begin{proposition}
  The invertible diffusion-denoising process is an asymmetric interpolation.
\end{proposition}

\subsection{Diffusion-based Out-of-distribution Detection}
\label{sec_main_method}

In the above, we show the advantages of DDP using a toy example and theoretical analysis. Here, we show the effect of DDP in a more complex example and design proper indicator scores. Specifically, we consider the last convolutional layer of a multi-layer model. In this setting, the input of this layer has many channels. To be more precise, the input is in $\mathbb{R}^{m\times m \times c}$ and is a combination of $c$ elements in the feature space $\mathbb{R}^{m\times m}$. As a result, a single input becomes $c$ points in the feature space instead of a single point.

\begin{figure}[t]
  \begin{minipage}{.49\linewidth}
    \centering
    \includegraphics[width=\linewidth, trim=200 120 340 60]{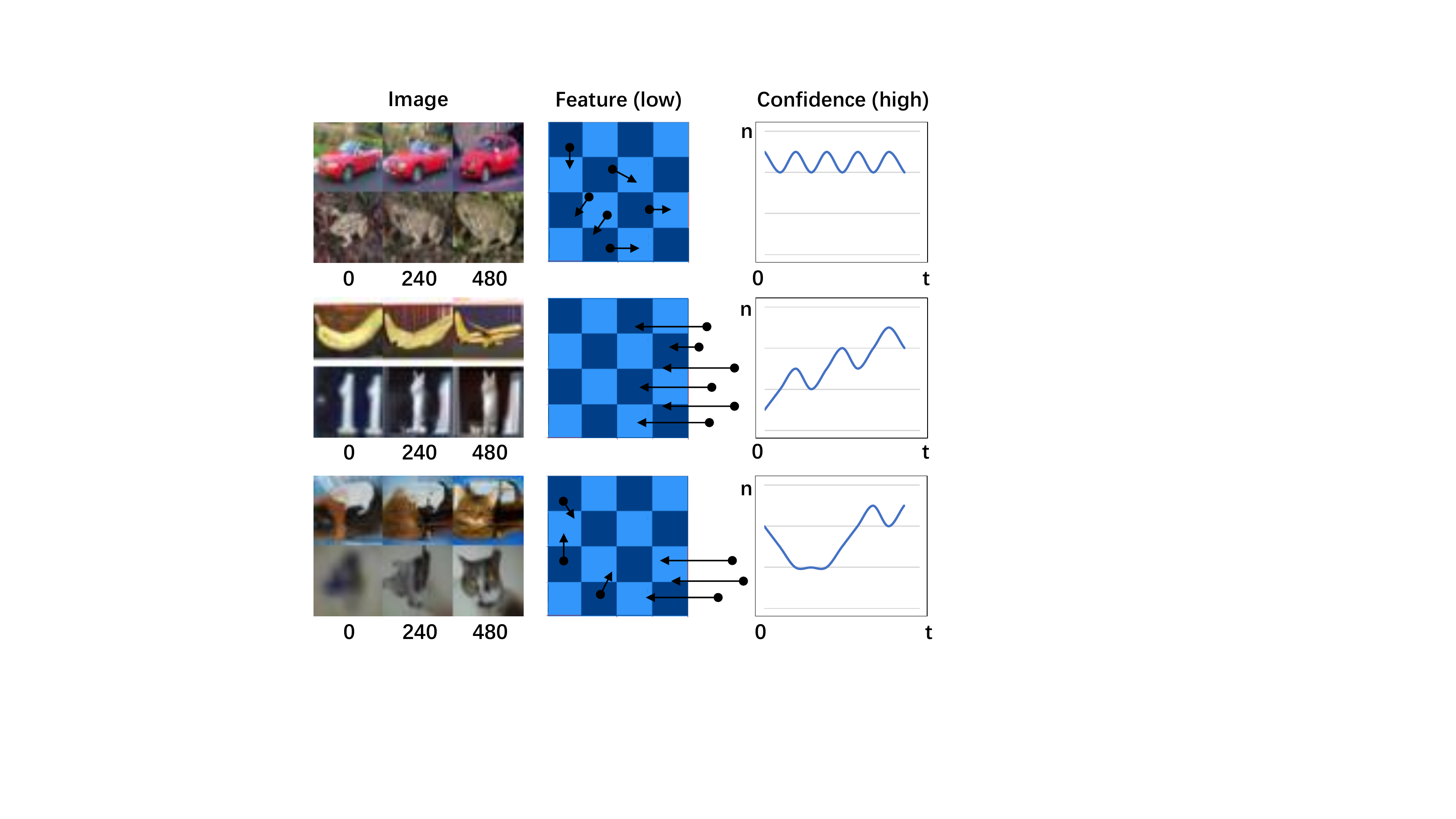}
    \caption{The change in image space and different feature spaces under the disturbance of DDP.}
    \label{fig_ddp_visual}
  \end{minipage}
  \ \ 
  \begin{minipage}{.49\linewidth}
    \vspace*{-0.5cm}
    \begin{algorithm}[H]
      \caption{Conditional Reconstruction}
      \label{alg_3}
      \begin{algorithmic}[1]
          \State {\bfseries Input:} Image $x$, timestep $t$
          \State $y = \text{FC}(\text{ResNet}(x))$
          \State $\epsilon \sim \mathcal{N}(0, 1)$
          \State $x_{\text{noise}} = \sqrt{\bar{\alpha}_t}x + \sqrt{1 - \bar{\alpha}_t}\epsilon$
          \State $x_{\text{recon}} = \Phi(x_{\text{noise}}, y, t, 0)$
          \State {\bfseries return} $x_{\text{recon}}$
      \end{algorithmic}
  \end{algorithm}
  \vspace*{-0.65cm}
  \begin{algorithm}[H]
    \caption{OOD detection}
    \label{alg_4}
    \begin{algorithmic}[1]
        \State {\bfseries Input:} Image $x$, additional samples $x_{\text{recon}}$
        \State fea = ResNet($x$)
        \State $\text{fea}_{\text{recon}} = \text{ResNet}(x_{\text{recon}})$
        \State score = $\sum|\text{fea} - \text{fea}_{\text{recon}}|$
        \If{$\text{score} > \delta$}
            \State {\bfseries return} OOD
        \EndIf
    \end{algorithmic}
  \end{algorithm}
  \end{minipage}
    
  \vspace*{-0.5cm}
\end{figure}

In Figure \ref{fig_ddp_visual}, we abstract the feature space into 2-dimensional space, and the ideal feature space related to the target class is the light blue square. The points in the feature space represent $c=6$ features of a single input, and the corresponding arrow shows the information flow of these features under DDP. We use the dark blue area to show the features used by the convolutional kernels of the last convolutional layer. The output of each convolutional kernel is a single value, the number of points $N$ that fall in its sensitive area. Then, we sum them to get the final result. This example is a simplification of the process used to compute confidence. %

For a normal InD input (the top row), DDP changes the image but still maintains it in the InD area. Its corresponding feature points move within the class-related feature area (light blue area). The number of feature points that can be captured by the discriminator (dark blue area) maintains a dynamic balance. For a normal OOD input (the middle row), the image contains no class-related features. Its feature points fall outside the class-related feature area. DDP pulls the OOD input to the high-density area of the training data. For example, a banana becomes a yellow plane, and the number one becomes a white cat. The number $N$ of the captured feature points increases at the same time. For more challenging OOD data (the bottom row), it is not InD, but it contains some class-related features. For example, a furry toy and the number four are incorrectly identified as cats. Its class-related features are relatively sparse but happen to be captured by the discriminator (dark blue area) by coincidence or man-made. However, the imbalance distribution of points in the dark blue area and the remaining causes more points to go out than come in under DDP, which causes a rapid decline of $N$. Therefore, all two kinds of OOD input can be detected by the change of confidence. Our main advantage is that we detect the balance between the features of interest to the discriminator and the remaining class-related features. This balance ensures that we can indirectly understand the whole feature space by detecting only the features of interest to the discriminator. More analysis and results about the effect of DDP are in Appendix \ref{sec_app_e_ddp}.

\textbf{Multiclass Detection} In real cases, it is often necessary to consider multiple categories. One key change is that the output is the probability of multiple categories, which is a multi-value vector. As a result, we need to modify our object of detection from the change of single-value confidence to a multi-value vector. This inspires us that the form of output is not limited, and we can also detect changes in any intermediate feature space as our indicator score.

The second change is that we need to not only keep the inputs within the InD area but also prevent category migration for InD inputs. In this case, we require the interpolation property, and there are two options available. Firstly, we can search for the nearest neighbor of the input in the image space and generate corresponding noises. We can then interpolate these noises with the original input using DDP. Another more intriguing option is to train a conditional diffusion model and fix the class condition to the class of the input.\footnote{When the label is unavailable, we can generate a pseudo-label using the discriminator model.} Then, all noise corresponds to images in the target class. Instead of searching for noise first, we can interpolate the input with any noise now.

In summary, our method uses conditional and invertible DDP to facilitate the flow of information and uses a discriminator model to generate the features. We then determine the OOD samples based on the dynamic change in the feature spaces. Our algorithms are in Algorithm \ref{alg_3} and \ref{alg_4}.

\section{Experiment}
\label{sec_exp}

\subsection{Setup}

We evaluate our methods on OpenOOD benchmarks \citep{yang2022openood}, using CIFAR10, CIFAR100, and ImageNet as InD datasets. For CIFAR10, we use CIFAR100, TinyImagenet \citep{krizhevsky2017imagenet}, SVHN \citep{netzer2011reading}, Texture \citep{cimpoi2014describing} and Places365 \citep{zhou2017places} as OOD datasets. For CIFAR100, the OOD datasets are the same, except for swapping CIFAR100 for CIFAR10 as the OOD dataset. For ImageNet, we use Species \citep{hendrycks2019scaling}, iNaturalist \citep{huang2021mos}, Texture, Places365, and SUN \citep{xiao2010sun} as OOD datasets. To ensure that there is no overlap between the InD and OOD datasets, all OOD datasets are meticulously filtered. 

For a fair comparison, we first train discriminator and generation models using the training set. We evaluate the results by calculating FPR@95 and AUROC \cite{fawcett2006introduction} between the test set of the InD dataset and others, which avoids the influence of model overfitting. All images from different datasets are resized into $32\times 32$ for CIFAR10/100 and $224\times 224$ for ImageNet. For CIFAR10/100, we utilize a pre-trained ResNet18 \citep{he2016deep} and the diffusion model from DDPM, trained by ourselves. For ImageNet, we use a pre-trained ResNet50 and a pre-trained latent diffusion model \citep{rombach2022high}. More detailed settings and hyperparameters of our method can be found in Appendix \ref{app_exp_detail}.

\begin{table}[t]
  \vspace*{-0.2cm}
  \centering
  \small
  \caption{The AUROC results of different methods on CIFAR10 and CIFAR100. The higher results are better and the bold results are the best in each case.} %
  \vspace*{0.2cm}
  \label{tb_ood}
  \begin{tabular}{@{}l|ccccc|ccccc@{}}
      \toprule
      InD & \multicolumn{5}{c|}{cifar10} & \multicolumn{5}{c}{cifar100} \\
      OOD & cifar100 & tin & svhn & texture & place & cifar10 & tin & svhn & texture & place \\ \midrule
      ODIN & 77.76 & 79.65 & 73.41 & 80.76 & 82.61 & 78.1 & 81.33 & 70.97 & 79.31 & 79.76 \\
      EBO & 86.19 & 88.61 & 88.42 & 86.88 & 89.62 & 79.07 & 82.46 & 77.81 & 77.84 & 80.16 \\
      ReAct & 86.37 & 88.91 & 89.52 & 88.19 & 90.1 & 73.48 & 79.63 & 84.45 & 83.58 & 76.94 \\
      MLS & 86.14 & 88.53 & 88.47 & 86.89 & 89.5 & \textbf{79.18} & 82.59 & 77.68 & 77.94 & \textbf{80.29} \\
      VIM & 87.19 & 88.86 & \textbf{97.28} & \textbf{96.03} & 90.03 & 71.54 & 78.34 & 81.15 & \textbf{87.41} & 75.77 \\
      KNN & 89.62 & 91.48 & 95.07 & 92.84 & 91.86 & 76.48 & 83.33 & 82.09 & 83.69 & 79.03 \\ \midrule
      Diff(ours) & \textbf{90.53} & \textbf{92.85} & 95.09 & 93.66 & \textbf{92.65} & 76.43 & \textbf{84.23} & \textbf{84.96} & 80.64 & 78.7 \\ \bottomrule
  \end{tabular}
  \vspace*{-0.4cm}
\end{table}

\begin{table}[t]
  \centering
  \small
  \caption{The AUROC results of different methods on ImageNet and the mean and standard deviation of the results on CIFAR10, CIFAR100, and ImageNet. A higher mean value is better, while a lower standard variance is better. The best results are in bold.}
  \label{tb_imagenet}
  \vspace*{0.2cm}
  \begin{tabular}{@{}l|ccccc|cccccc@{}}
    \toprule
    InD & \multicolumn{5}{c|}{imagenet} & \multicolumn{2}{c}{cifar10} & \multicolumn{2}{c}{cifar100} & \multicolumn{2}{c}{imagenet} \\
    OOD/stat & specie & inatur & texture & place & sun & mean & std & mean & std & mean & std \\ \midrule
    ODIN & 71.51 & 91.15 & 87.5 & 84.5 & 86.9 & 78.8 & 3.14 & 77.9 & 3.61 & 84.3 & 6.75 \\
    EBO & 71.98 & 90.63 & 86.73 & 83.97 & 86.58 & 87.9 & 1.24 & 79.5 & \textbf{1.73} & 84.0 & 6.36 \\
    ReAct & 77.43 & 96.4 & 90.49 & \textbf{91.92} & \textbf{94.4} & 88.6 & 1.29 & 79.6 & 4.10 & 90.1 & 6.67 \\
    MLS & 72.89 & 91.15 & 86.41 & 84.19 & 86.6 & 87.9 & \textbf{1.22} & 79.5 & 1.79 & 84.2 & 6.11 \\
    VIM & 69.43 & 86.4 & \textbf{97.35} & 76.99 & 79.62 & 91.9 & 4.02 & 78.8 & 5.32 & 82.0 & 9.42 \\
    KNN & 66.84 & 84.16 & 96.02 & 70.22 & 76.65 & 92.2 & 1.79 & 80.9 & 2.76 & 78.8 & 10.5 \\ \midrule
    Diff(ours) & \textbf{85.03} & \textbf{97.05} & 94.76 & 88.86 & 92.99 & \textbf{93.0} & 1.49 & \textbf{81.0} & 3.24 & \textbf{91.7} & \textbf{4.29} \\ \bottomrule
    \end{tabular}
  \vspace*{-0.2cm}
\end{table}

\subsection{Out-of-distribution Detection}

In Table \ref{tb_ood}, we choose six representative baselines, which do not adjust the discriminator model similar to our method. ODIN \citep{liang2018enhancing} uses temperature scaling and gradient-based input perturbation. EBO \citep{liu2020energy} uses an energy-based function. ReAct \citep{sun2021react} uses rectified activation. MLS \citep{hendrycks2019scaling} uses maximum logit scores. VIM \citep{wang2022vim} combines the information of feature space and logic space. KNN \citep{sun2022out} uses the nearest neighbor in the feature space. Our method surpasses all these methods in five out of ten cases and achieves comparable results in the remaining cases. In Table \ref{tb_imagenet}, we find that our method performs better than the other methods in two out of five cases and achieves comparable results in the remaining cases. This shows the effectiveness of our method on datasets of varying sizes. Moreover, our method has low standard variance across different OOD datasets, indicating its excellent generalization ability. FPR@95 results can be found in Appendix \ref{app_fpr}.

\begin{wraptable}{r}{0.6\linewidth}
    \vspace*{-0.45cm}
    \small
    \centering
    \caption{The AUROC results when we treat misclassified data in CIFAR100 as a kind of OOD data.}
    \label{tb_new_exp}
    \vspace*{-0.1cm}
    \begin{tabular}{@{}l|cccccl@{}}
        \toprule
        OOD & wrong & cifar10 & tin & svhn & texture & place \\ \midrule
        MLS & 83.93 & \textbf{86.50} & 89.30 & 85.16 & 85.49 & 87.55 \\
        VIM & 79.31 & 78.03 & 84.26 & 87.00 & \textbf{91.19} & 81.96 \\
        KNN & 84.36 & 83.59 & 89.88 & 88.98 & 89.75 & 86.19 \\ \midrule
        Diff & \textbf{84.96} & 85.67 & \textbf{91.87} & \textbf{96.50} & 89.56 & \textbf{88.27} \\ \bottomrule
    \end{tabular}
    \vspace*{-0.5cm}
\end{wraptable}

OOD detection is widely used in scenarios with high-security requirements. It is also imperative to identify any misclassified data from the InD dataset. Thus, we treat the misclassified data of the pre-trained ResNet18 as a new form of OOD data. ResNet18 has an accuracy of only 78\% on CIFAR100, making it an appropriate candidate for demonstrating the impact of misclassified data. Our method achieves SOTA results in distinguishing between correctly and incorrectly classified data in the training datasets. Under this setting, MLS and VIM outperform us on only one OOD dataset, respectively, and the gap is small. On all other datasets, our method is substantially ahead of all other methods. By comparing the results in the corresponding positions in Table \ref{tb_ood} and \ref{tb_new_exp}, we find that after removing the misclassified data from InD, the OOD detection results of all methods are improved, with our method having the biggest improvement. This means that all methods tend to treat misclassified data as OOD data. The detection results of misclassified data and classical OOD data are mixed, which limits the classical OOD detection results. 

\subsection{Ablation Study}
\label{sec_abla}

Here, we show the influence of each item and hyperparameter in our method. Here, the AUROC results are the average of five AUROC results using all OOD datasets on CIFAR10. More ablation study results can be found in Appendix \ref{app_ablation}.

\begin{figure*}[tp]
    \begin{minipage}[t]{0.245\linewidth}
      \centering
      \includegraphics[width=\linewidth, trim=10 -5 40 40]{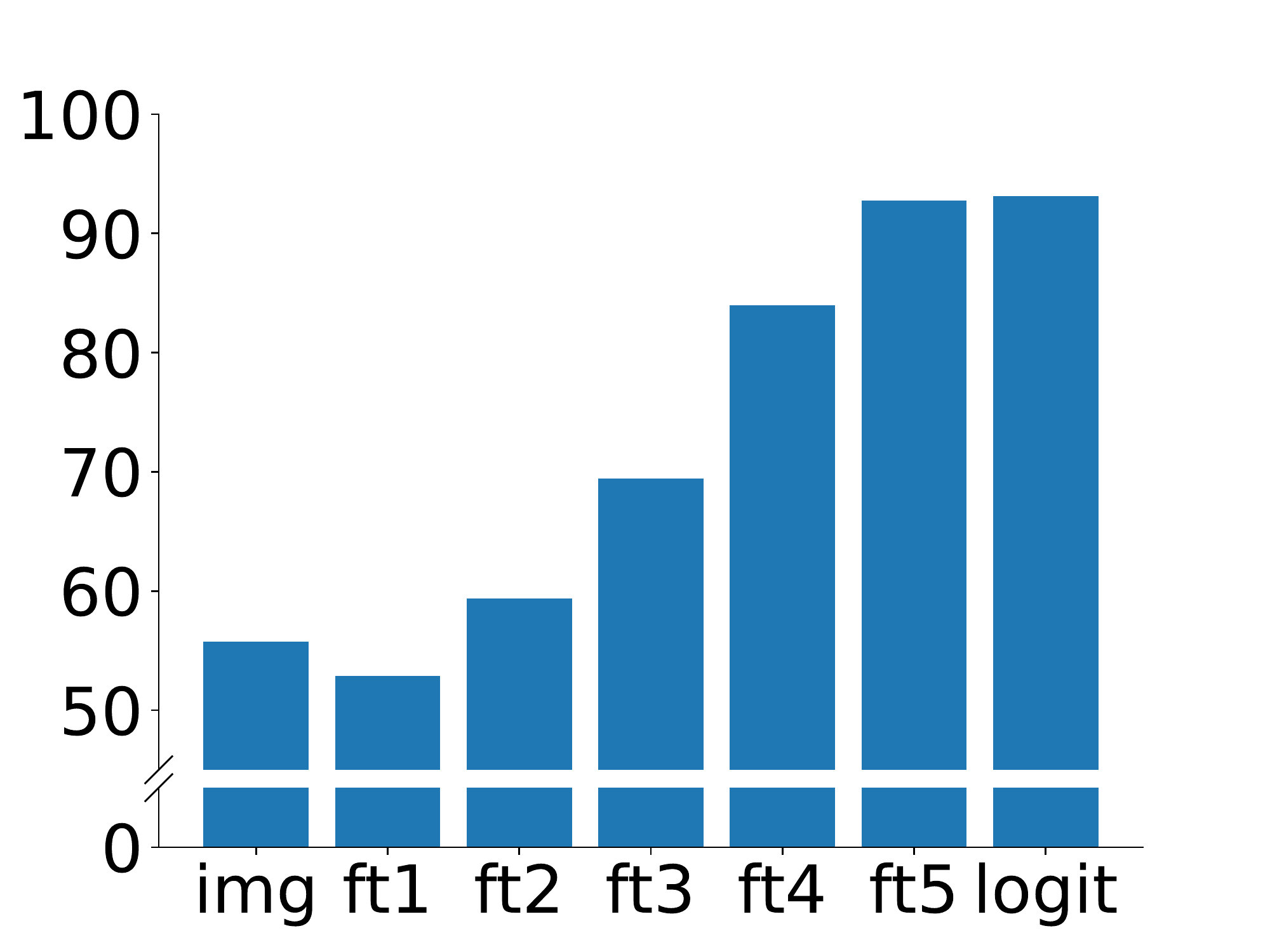}
      \vspace*{-0.55cm}
      \label{fig_space}
  \end{minipage}
  \begin{minipage}[t]{0.245\linewidth}
        \centering
        \includegraphics[width=\textwidth, trim=5 5 0 0, clip]{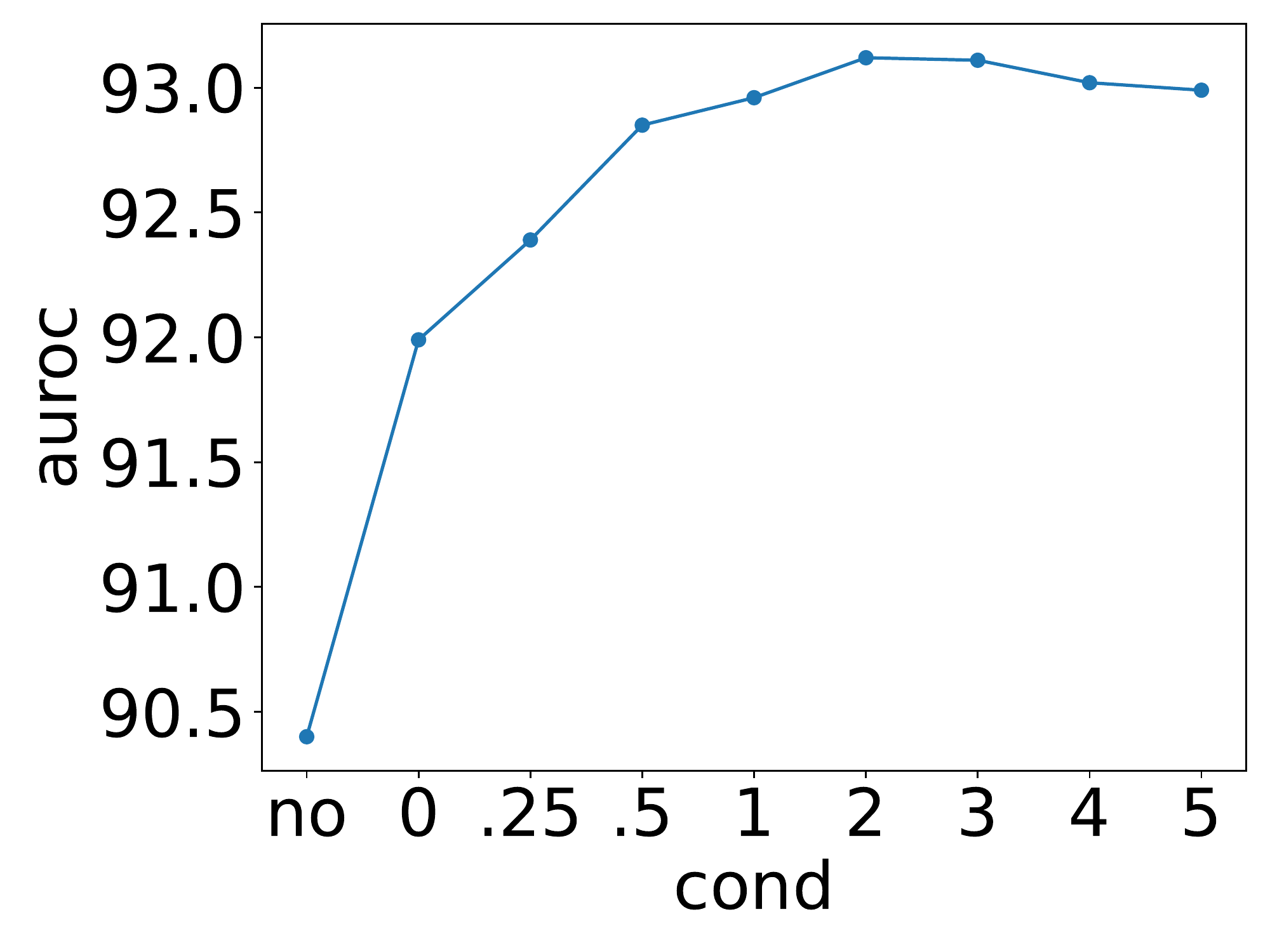}   
        \vspace*{-0.55cm}
        \label{fig_cond}
      \end{minipage}
      \begin{minipage}[t]{0.245\linewidth}
        \centering
        \includegraphics[width=\textwidth, trim=5 5 0 0, clip]{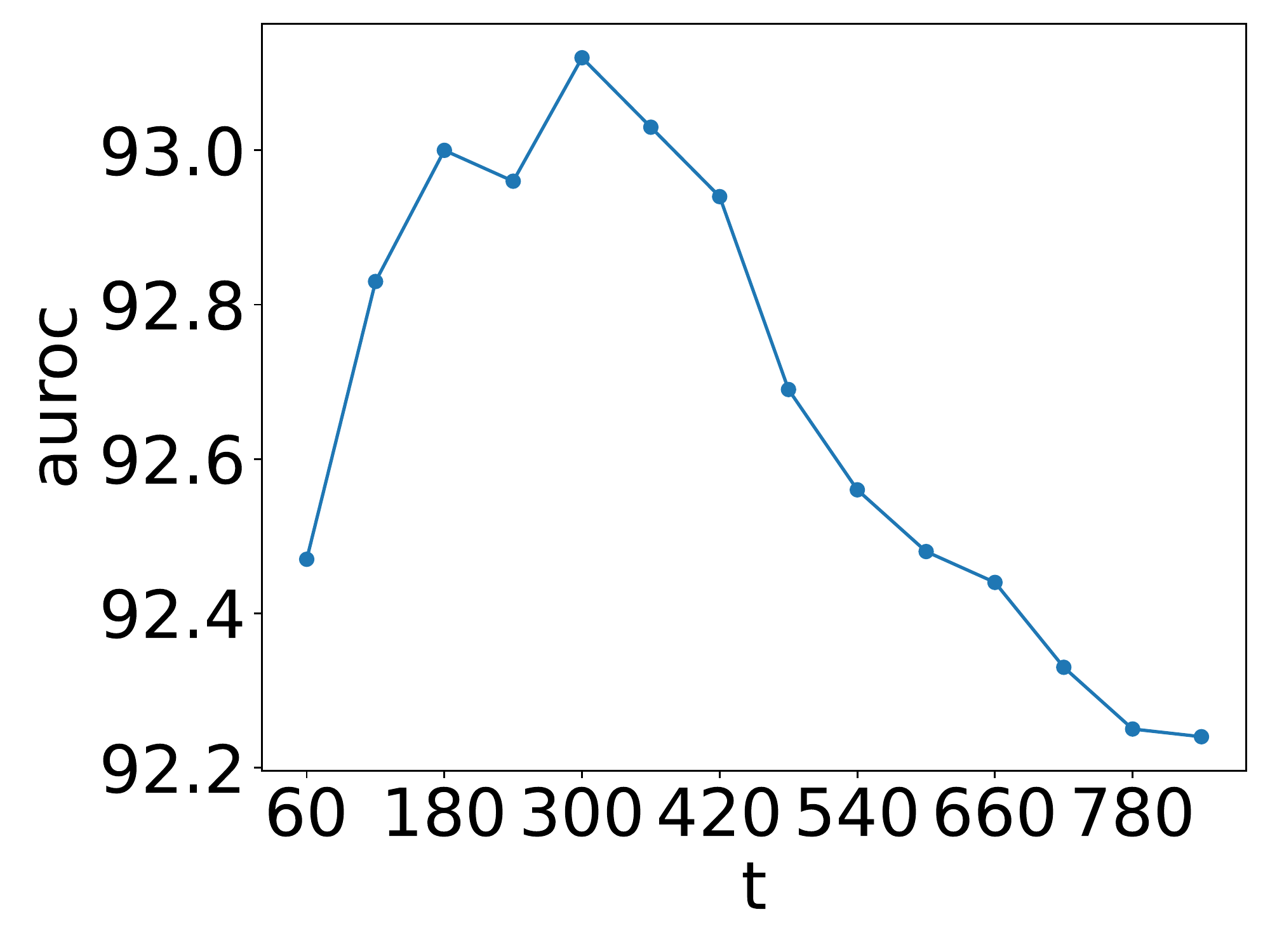}  
        \vspace*{-0.55cm}
        \label{fig_timestep}
      \end{minipage}
      \begin{minipage}[t]{0.245\linewidth}
        \centering
        \includegraphics[width=\textwidth, trim=5 5 0 0, clip]{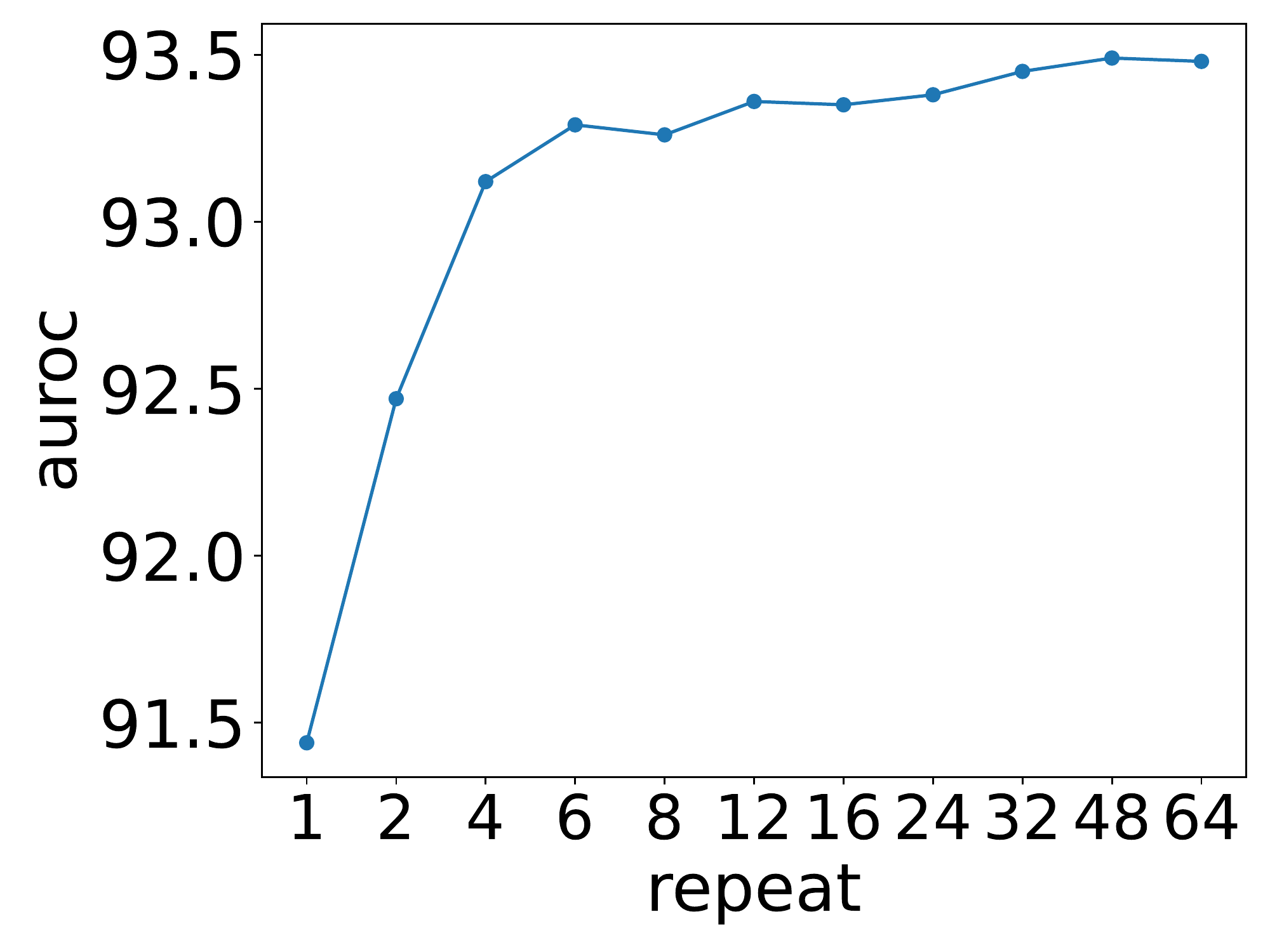} 
        \vspace*{-0.55cm}  
        \label{fig_repeat}
    \end{minipage}
    \vspace*{-0.4cm}
    \caption{the AUROC results obtained under different detection spaces, guidance weights, disturbance degrees, and resampling sizes on CIFAR10.}
    \label{fig_ablation}
    \vspace*{-0.5cm}
\end{figure*}

\textbf{Detection space} In Figure \ref{fig_ablation}.1, we compare the results obtained when using different detection spaces, including the input image space, different level feature spaces, and the logit space. We achieve the best results when using the high-level feature or the logit. The low-level features and image space yield relatively poor results, primarily due to the redundancy of information at these levels. This highlights the importance of the discriminator model to the generative model in OOD detection. We also observe that setting a threshold for the final feature space can improve detection results, similar to ReAct \citep{sun2021react}.

\textbf{Condition} In Figure \ref{fig_ablation}.2, we examine the impact of different guidance weights $\omega$ on conditional diffusion models. We find that higher guidance weights result in relatively better outcomes, indicating that realness is more critical than diversity in the OOD detection task. Furthermore, we find that the conditional version outperforms the unconditional version, with the detection results showing a more significant improvement on CIFAR100. This suggests that conditional control is essential, and higher guidance weights $\omega$ are required for more classes.

\textbf{Timestep} In Figure \ref{fig_ablation}.3, we investigate how to select the optimal value of $t$ in DDP. We find that the best choice is $t=300$ for small datasets. When $t\leq 300$, the difference caused by DDP is still not distinct enough. When $t\geq 300$, the noise item accounts for a larger and larger proportion, causing information loss and limiting the OOD detection results. For large datasets, the information in images is redundant, necessitating more noise to create a noticeable change. The best $t$ is 600 in this case.

\textbf{Resampling} In Figure \ref{fig_ablation}.4, we determine the influence of the repeated sampling size. According to our analysis, the consistent detection results are maintained by dynamic balance. Therefore, we need to resample several times to eliminate the random error in DDP. We observe that a larger number of resampling leads to better results, and four times resampling is sufficient.

\section{Conclusion}

In this paper, we start with the analysis that discriminator models may ignore common features which are important for OOD detection. Based on that, we provide a new perceptron bias assumption to explain the overconfidence problem. Under this assumption, we first show the potential of generation models and develop the diffusion denoising process of diffusion models as the solution. Then, we demonstrate how to design indicator scores in real-world scenarios and how to apply our method to multi-class situations. Our approach is highly interpretable and employs simple training objectives, including softmax loss for discriminator models and L2 loss for diffusion models. Using the combination of a ResNet and a diffusion model, we get SOTA results on CIFAR10, CIFAR100, and ImageNet. We believe that our strategy can benefit the OOD detection community and has the potential to be applied to OOD generalization and other related tasks.

While the limitations of our method include the detection speed and the training cost of diffusion models, the advancements in effective training and sampling for diffusion models are rapidly progressing. For instance, consistency models \citep{song2023consistency} can complete sampling in several steps, and ControlNet \citep{zhang2023adding} can improve training efficiency. We aim to leverage these new developments to enhance the efficiency of our method in the future.

\begin{ack}
  We would like to thank Hao Ying and Ziyue Jiang from Zhejiang University for their insightful discussions during the course of this paper. We would especially like to thank Chongxuan Li from the Renmin University of China. Chongxuan Li provides the idea of conditional diffusion models is essential for this paper.
\end{ack}

{
  \small
  \bibliography{neurips_2023}
  \bibliographystyle{unsrtnat}
}

\newpage
\appendix

\section{Supplementary Material}

\subsection{Related Work}
\label{app_related}

Here, we introduce more related works to show the recent development of diffusion models. Then we show similarities and differences between our method and existing works to emphasize our advantages.

\subsubsection{Diffusion Model}

Denoising Diffusion Probabilistic Models (DDPMs) \cite{ho2020denoising} successfully generate high-quality images and make DMs become popular. For now, DMs can not only generate unconditional high-quality images but also are applied to many different fields. For the conditional generation task, DMs can do interpolation, manipulation, image-editing, style transformation and text-conditional generation \cite{ramesh2021zero, ramesh2022hierarchical}. For different data types, DMs can do text \cite{austin2021structured}, audio \cite{kong2020diffwave, lam2021bddm} and video generation. 

The main challenge for DMs is that they require hundreds to thousands of iterations to produce high-quality results, which limits the application of DMs. After DDPMs, many works try to make DMs faster and better. Some of them focus on the denoising equations of DMs. \citet{nichol2021improved} design a better time schedule for the denoising process. \citet{liu2022pseudo} provide new numerical methods for the denoising process. \citet{bao2022analytic} find analytic results for the variance of the denoising process. Some of them try to design new training strategies and new models. \citet{salimans2021progressive} use distillation to accelerate DMs. \citet{dhariwal2021diffusion} change the model structure from Unet to GAN to make each iteration step more powerful. What's more, \citet{song2021scorebased} first shows that DMs can be rewritten as two neural differential equations \cite{chen2018neural, dupont2019augmented}. Therefore, the numerical methods used to accelerate neural differential equations can also be used here. 

\subsubsection{Similarity and difference}

Several baselines are similar to our method in some ways. The first one is KNN \citep{sun2022out}, which does a KNN search in the feature space. However, KNN ignores the possibility that an OOD input may have a similar feature as InD data. All methods that only use the final feature have this problem. They can not solve the information loss of discriminator models. In addition, using KNN in the input space directly is also invalid, because of data sparsity and irrelevant information interference. Our method uses ResNet to extract useful information and uses diffusion models to solve the information loss at the same time.

Another similar approach is data generation. Existing methods use generation models to generate OOD data \citep{ge2017generative, neal2018open, marek2021oodgan, du2022vos} to help the discriminator models to know the capability boundary. This requires careful design of additional loss functions and the whole pipeline. Our method uses classical loss functions and training processes for both discriminator and generation models. Then our method uses the generation models to do interpolation between the new input data and the training data and puts all results into the discriminator model.

Data augmentation is also similar to our method. Some of them use augmentation results to retrain discriminator models, but our generation and discriminator models are trained separately. Although these methods can enhance the richness of data and keep the data realistic, some new methods start to add complex and unreal augmentation \citep{hendrycks2022pixmix}, which increases the burden on the models and lacks clear motivation. Some of them only use augmentation results in the detection processes. However, existing methods focus on the denoising process, which is a trade-off between remaining InD data unchanged and enhancing the input, which becomes an obvious bottleneck. We find that keeping InD related unchanged is unnecessary. Our method replaces the denoising operator with an interpolation operator. The interpolation operator only remains the InD input in the InD area, which is good enough for OOD detection and provides more possibilities to enhance the input.

\subsection{Invertible Diffusion Model}
\label{app_inv_diff}

Here, we introduce more details of the invertible diffusion model used in this paper. Then, we analyze how to make the iteration of diffusion models more invertible in practice. After that, we introduce asymmetric and symmetric interpolations under the invertible condition.

\subsubsection{Method detail}

\paragraph{Score-based generation model}

\citet{song2021scorebased} show that the diffusion-denoising process can also be treated as two differential equations:
\begin{equation}
  \begin{split}
    dx &= (\sqrt{1 - \beta(t) - 1})x(t)dt + \sqrt{\beta(t)}dw \\
    dx &= \left((\sqrt{1 - \beta(t) - 1})x(t) - \frac{1}{2}\beta(t)s_\theta(x(t), t)\right)dt.
  \end{split}
\end{equation}
This is called probability flows (PFs). The noise $\epsilon_\theta$ of DMs and the gradient of logic likelihood $s_\theta$ are equivalent \cite{bao2022analytic}. More specifically, we have that $s_\theta(x, t) = -\frac{1}{1-\bar{\alpha}_t}\epsilon_\theta(x, t)$.

\paragraph{Pseudo numerical method}

\citet{liu2022pseudo} provide pseudo numerical methods for diffusion models (PNDMs) to accelerate DDIMs. PNDMs define Equation (\ref{eq_ddpm}) with $\sigma_t=0$ as transfer function:
\begin{equation}
  \phi(x_t, \epsilon_t, t, t-\delta) = \frac{\sqrt{\bar{\alpha}_{t-\delta}}}{\sqrt{\bar{\alpha}_t}}x_t - 
  \frac{(\bar{\alpha}_{t-\delta}-\bar{\alpha}_t)}{\sqrt{\bar{\alpha}_t}(\sqrt{(1-\bar{\alpha}_{t-\delta})\bar{\alpha}_{t}} + \sqrt{(1-\bar{\alpha}_{t})\bar{\alpha}_{t-\delta}})}\epsilon_t.
  \label{eq_transfor}
\end{equation}
PNDMs combine this transfer function with the noise estimated by classical numerical methods, like the linear multistep method, to get the new denoising equations:
\begin{equation}
  \begin{cases}
     &\epsilon_t' = \frac{1}{24}(55\epsilon_t-59\epsilon_{t+\delta}+37\epsilon_{t+2\delta}-9\epsilon_{t+3\delta}) \\
     &x_{t-\delta} = \phi(x_t, \epsilon_t', t, t-\delta).
  \end{cases}
  \label{eq_lms}
\end{equation}
Here, $\epsilon_t = \epsilon_\theta(x_t, t)$. Both PFs and PNDMs accelerate the denoising process without loss of quality.

\subsubsection{Invertibility} 

\begin{wrapfigure}{r}{0.3\linewidth}
    \vspace*{-0.35cm}
    \centering
    \includegraphics[width=\linewidth, trim=15 20 40 40]{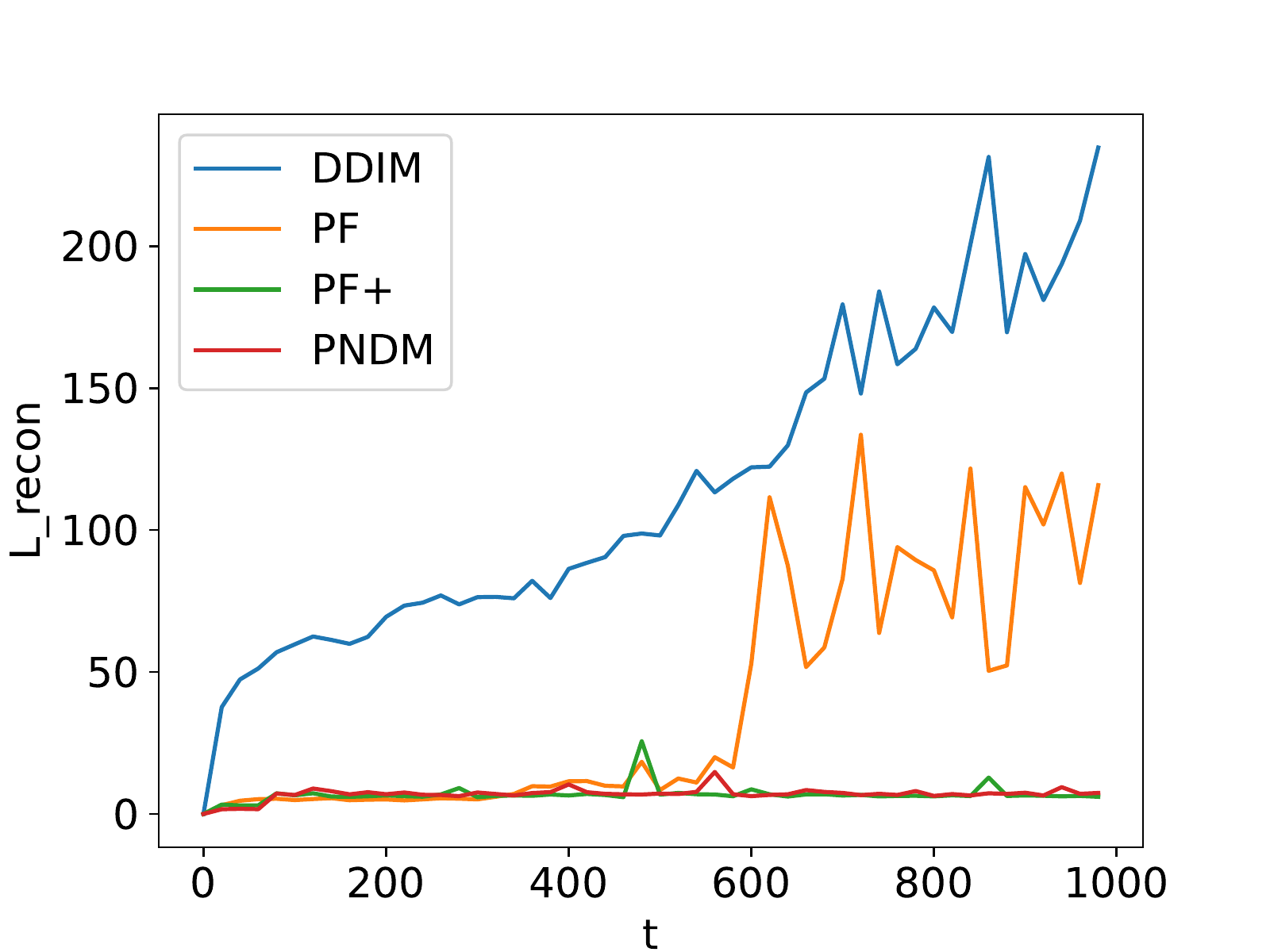}
    \caption{The reconstruction error under different iteration interval $[0, t]$ and fixed step size 20.}
    \label{fig_inv_l1}
    \vspace*{-0.3cm}
  \end{wrapfigure}

We show the test results in Figure \ref{fig_inv_l1}. For DDIMs, the error occurs at the beginning, and the error accumulates with the increase of the total generation step. For PFs, the initial error is not huge, but the cumulative error occurs when the number of the total generation steps is bigger than 500. DDIMs are first-order methods, and other methods are high-order methods. We can say higher convergent order can increase the invertibility. PFs use numerical methods of adaptive step size, and PNDMs use methods of fixed step size. Therefore, we think that fixed step size can also benefit invertibility. To verify this, we replace the methods of adaptive step size used by PFs with the methods of fixed step size and call it probability flows plus (PFs+). We find that the errors decrease immediately. The reason is that fixed step size maintains consistency between the sampling locations of the forward and reverse processes, which benefits the invertibility. Combining the above analysis, we have the following property:  %

\begin{proposition}
  High convergent order and fixed iteration step size can improve the invertibility of DMs under fixed total iteration steps.
\end{proposition}

\subsubsection{Interpolation}
\label{app_itp}

\begin{figure}[htb]
  \centering
  \includegraphics[width=\linewidth, trim=160 160 205 120]{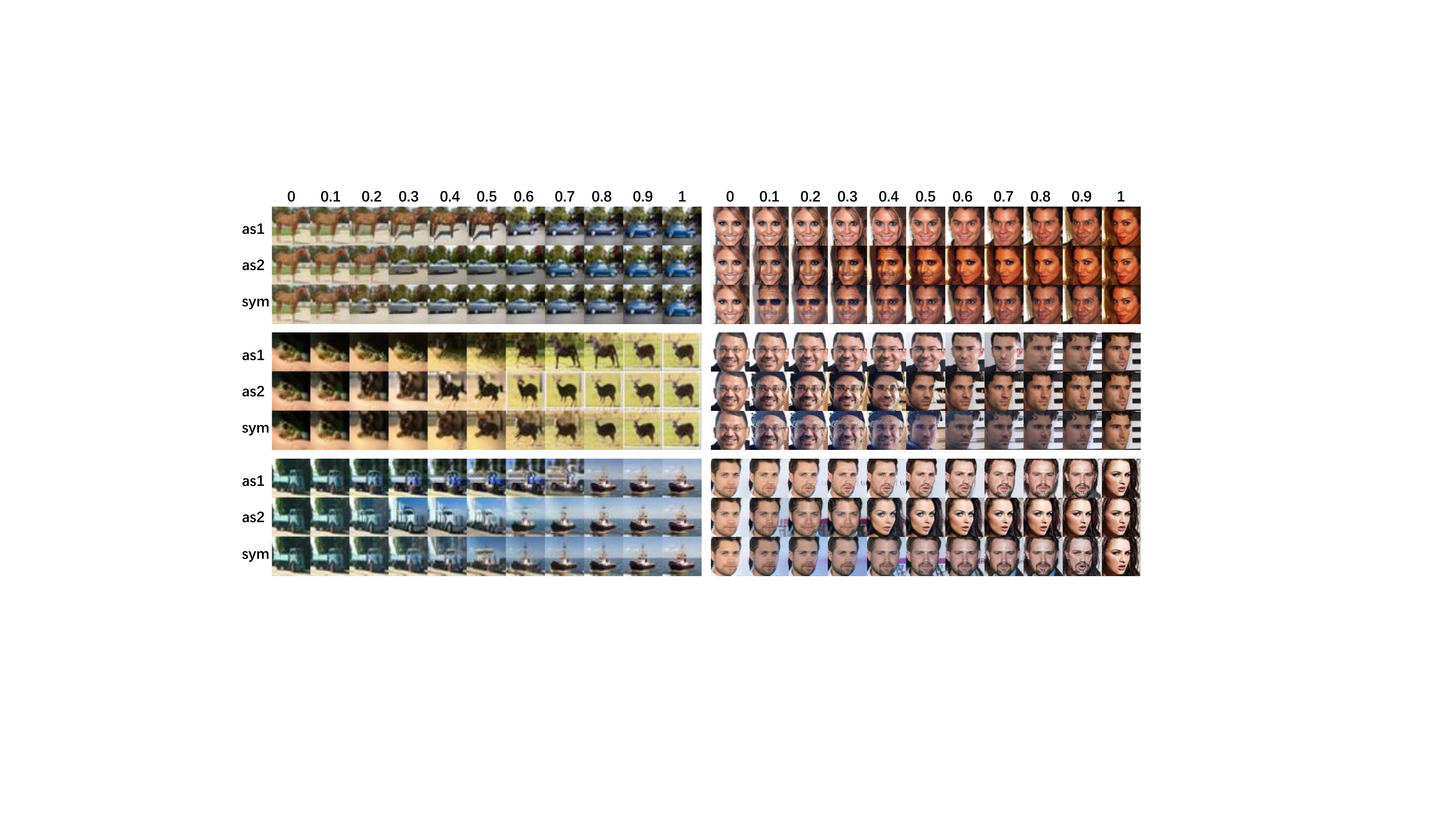}
  \caption{The results of asymmetric and symmetric interpolations using two fixed images from CIFAR10 or CelebA.}
  \label{fig_itp_app}
  \vspace*{-0.4cm}
\end{figure}

In Algorithm \ref{alg_itp11} and \ref{alg_itp22}, we introduce two types of interpolation using diffusion models. The positions of $x_1$ and $x_2$ are symmetric in the original interpolation and asymmetric for the asymmetric one. According to our experiment, $x_1$ is more important in asymmetric interpolation. The visualization of these interpolations can be found in Fig \ref{fig_itp_app}. 

Because the $x_1$ can be preserved at the beginning, which is more like a denoising operator. When the interpolation ratio becomes bigger, the results change from $x_1$ to $x_2$ smoothly, which is more like an interpolation operator. Therefore, DDP is a combination of denoising and interpolation operators and the timestep $t$ controls the ratio of denoising and interpolation.

\begin{figure}[h]
  \centering
  \begin{minipage}[t]{0.49\linewidth}
    \vspace*{-0.5cm}
    \begin{algorithm}[H]
      \caption{Symmetric interpolation}
      \label{alg_itp11}
      \begin{algorithmic}[1]
        \State {\bfseries Input:} Images $x^1, x^2$, generation gap $\delta$, interpolation rate $\sigma$
        \State $x_{T}^1, x_T^2 = \Phi(x^1, 0, T, \delta), \Phi(x^2, 0, T, \delta)$
        \State $x_T^{\text{inter}} = \text{Slerp}(x_{T}^1, x_T^2, \sigma)$
        \State $x^{\text{inter}} = \Phi(x_T^{\text{inter}}, T, 0, \delta)$ \\
        \State {\bfseries return} $x^{\text{inter}}$
      \end{algorithmic}
    \end{algorithm}
  \end{minipage}
  \ \ 
  \begin{minipage}[t]{0.49\linewidth}
    \vspace*{-0.5cm}
    \begin{algorithm}[H]
      \caption{Asymmetric interpolation}
      \label{alg_itp22}
      \begin{algorithmic}[1]
        \State {\bfseries Input:} Images $x^1, x^2$, generation gap $\delta$, interpolation timestep $t$
        \State $x_{T}^2 = \Phi(x^2, 0, T, \delta)$
        \State $x_t^{\text{inter}} = \sqrt{\bar{\alpha}_t} x^1 + \sqrt{1 - \bar{\alpha}_t} x_T^2$
        \State $x^{\text{inter}} = \Phi(x_t^{\text{inter}}, t, 0, \delta)$ \\
        \State {\bfseries return} $x^{\text{inter}}$
      \end{algorithmic}
    \end{algorithm}
  \end{minipage}
\end{figure}

\newpage
\subsection{Proof of toy example}
\label{app_proof}

Here, we prove that the moving and mixing operators can perfectly solve the toy example. To make this claim strict, we need to make some definitions. 

We first transfer OOD detection and the design of our two-step method into a formal problem. Following the definitions in Section \ref{sec_ood_bg}, the detector is a function $f: U\to \mathbb{R}$ that can be an analytic function or a neural network defined on a certain space $U$. The detection problem is that given a distribution $p$ on $U$, let $W \triangleq \{x\in U|p(x)>\sigma\}$ and we want to find a detector $f$ that satisfies $f^{-1}(1)=W$ exactly. The two-step method starts with a relatively simple detector $f$, which may treat bad cases as normal and $f^{-1}(1)$ is strictly bigger than $W$. Then we enhance the input to help the detection. We define a group of operators $\{g_i:U\to U\}_{i\in I_g}$\footnote{$I_g$, $I_S$ is the index set of $\{g_i\}$ and $\{S_i\}$ respectively. They can be finite or infinite.} satisfying $f\circ g_i(x)=f(x)=1$ if $x\in W$ and $\cap_{i\in I_g} (f\circ g_i)^{-1}(1) \subset f^{-1}(1)$. The first condition means that all $f\circ g_i$ can also identify the InD data correctly and the second condition means that we can minimize the gap between $\cap_{i\in I_g} (f\circ g_i)^{-1}(1)$ and $W$ by additional $\{g_i\}$. Now we can formally define our task as an \textbf{additional operator search} problem:

\begin{definition}
    Given a fixed subarea $W\subset U$, a single value function $f$ satisfying $W \subset f^{-1}(1)$, how can we design operators $\{g_i\}_{i\in I_g}$ to minimize $\cap_{i\in I_g} (f\circ g_i)^{-1}(1)$?
\end{definition}

Each image can be represented as a function $r$ on $[0, 1]^2$ and the value of $r(x, y)$ is the RGB value at position $(x, y)$. And the images are continuous in most positions. Therefore, we simplify the input space to $\mathcal{C}([0, 1]^2)$ the continuous function on $[0, 1]^2$ and then to one-dimensional $\mathcal{C}([0, 1])$ for simplicity. The mask operator is a restriction operator $\phi_S(r) = r|_{S=[0, 0.25]\cup[0.5, 0.75]}$ here. The InD is just $\{r\in \mathcal{C}([0, 1])\text{ }|\text{ } |r|\leq \sigma\}$ and $|r|$ is the max absolute value of $r$ on $[0, 1]$. The bad cases form a set that satisfies $\mathcal{A}_\sigma(\phi_S) = \{r| \phi_S(r)=\phi_S(0)=0, |r|>\sigma\}$. We can prove that $\mathcal{A}_\sigma(\phi_S)=\emptyset$ when we choose proper $\{g_i\}$.

A straightforward solution is that for each $r\in \mathcal{C}(\mathbb{R})$, let $\{g(x)_{i} = r(x+i)| i\in \{0, \pm 0.25\}\}$, which represents the moving operator. The proof is that we can use $g_0|_S$ to get the information about $r$ on $S$ and use $g_{\pm 0.25}|_S$ to get the remaining on $[0, 1]/S$. Then $r$ mush satisfies $|r(x)|<\sigma, \forall x\in [0,1]$. There also exist other kinds of solutions. For example, let $\{g(x)_{a, b} \equiv \frac{1}{b-a} \int_a^b r(x)dx| a, b\in [0, 1]\}$, which represents the mixing operator. In addition, the mask operator is not necessary and we can extend it to any restriction operator. %

Here we prove that $\{g(x)_{i} = f(x+i)| i\in \{0, \pm 0.25\}\}$ and $\{g(x)_{a, b} \equiv \frac{1}{b-a} \int_a^b f(x)dx| a, b\in [0, 1]\}$ can solve the information loss problem in the toy example.

For $\{g(x)_{i} = f(x+i)| i\in \{0, \pm 0.25\}\}$, $g_0|_S=f|_S$ and it equals to zero means that f is zero on S. And then we also know that $g_{0.25}|_S=f|_{S-0.25}$ and $g_{-0.25}|_S=f|_{S+0.25}$ equals zero, which means that $f$ is zero on $S-0.25\cup S+0.25 = [0, 1]/S$. Then the only choice of $f$ is zero and the annihilator set is empty.

For $\{g(x)_{a, b} \equiv \frac{1}{b-a} \int_a^b f(x)dx| a, b\in [0, 1]\}$. If $\exists x_0\in [0, 1]$, $f(x_0)> \delta$, then $\exists \epsilon$ s.t. $\forall x\in [x_0-\epsilon, x_0+\epsilon]$, $f(x)>\delta$ because $f$ is continuous. Then we have that $g(x)_{x_0-\epsilon, x_0+\epsilon} \equiv \frac{1}{b-a} \int_a^b f(x)dx > \delta$ and $g(x)_{x_0-\epsilon, x_0+\epsilon}|_S > \delta$, too. This does not satisfy our condition, so the annihilator set is empty.

\subsection{Effect of diffusion denoising process}
\label{sec_app_e_ddp}

Here, we show more detailed results to display the effect of the diffusion denoising process.

\subsubsection{Different level of feature}

\begin{figure}[htb]
  \vspace*{-0.4cm}
  \begin{minipage}[t]{0.33\linewidth}
    \centering
    \includegraphics[width=\textwidth, trim=8 -5.5 22 -1, clip]{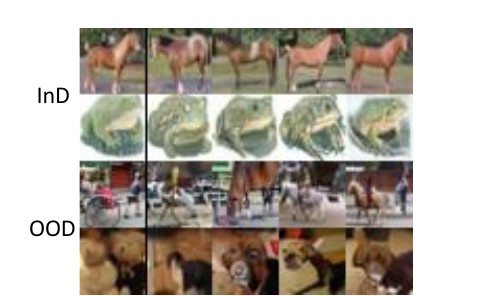}   
    \vspace*{-0.2cm}
 \end{minipage}
 \begin{minipage}[t]{0.33\linewidth}
    \centering
    \includegraphics[width=\textwidth, trim=0 0 40 30, clip]{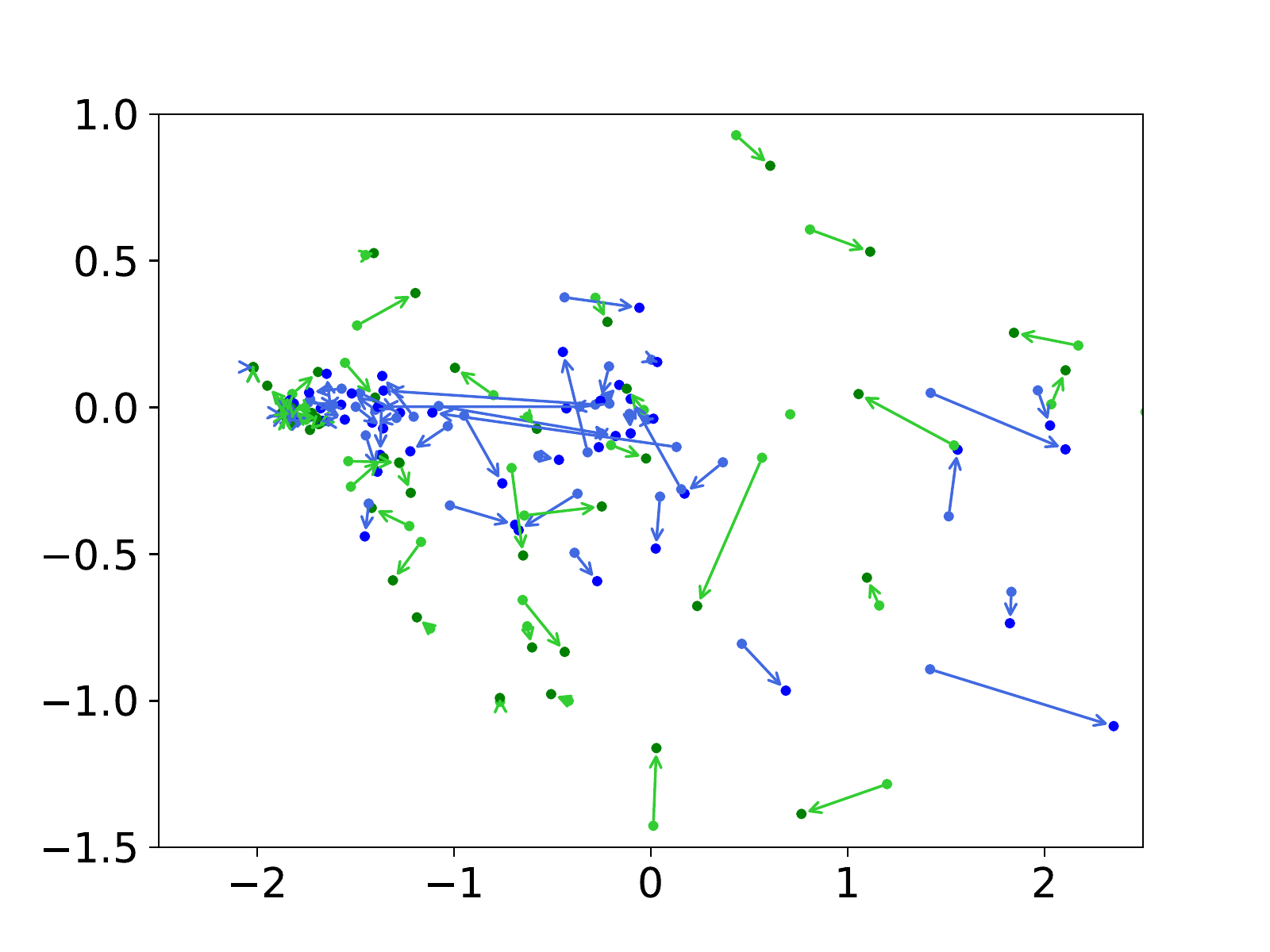}   
    \vspace*{-0.2cm}
  \end{minipage}
  \begin{minipage}[t]{0.33\linewidth}
    \centering
    \includegraphics[width=\textwidth, trim=0 0 40 30, clip]{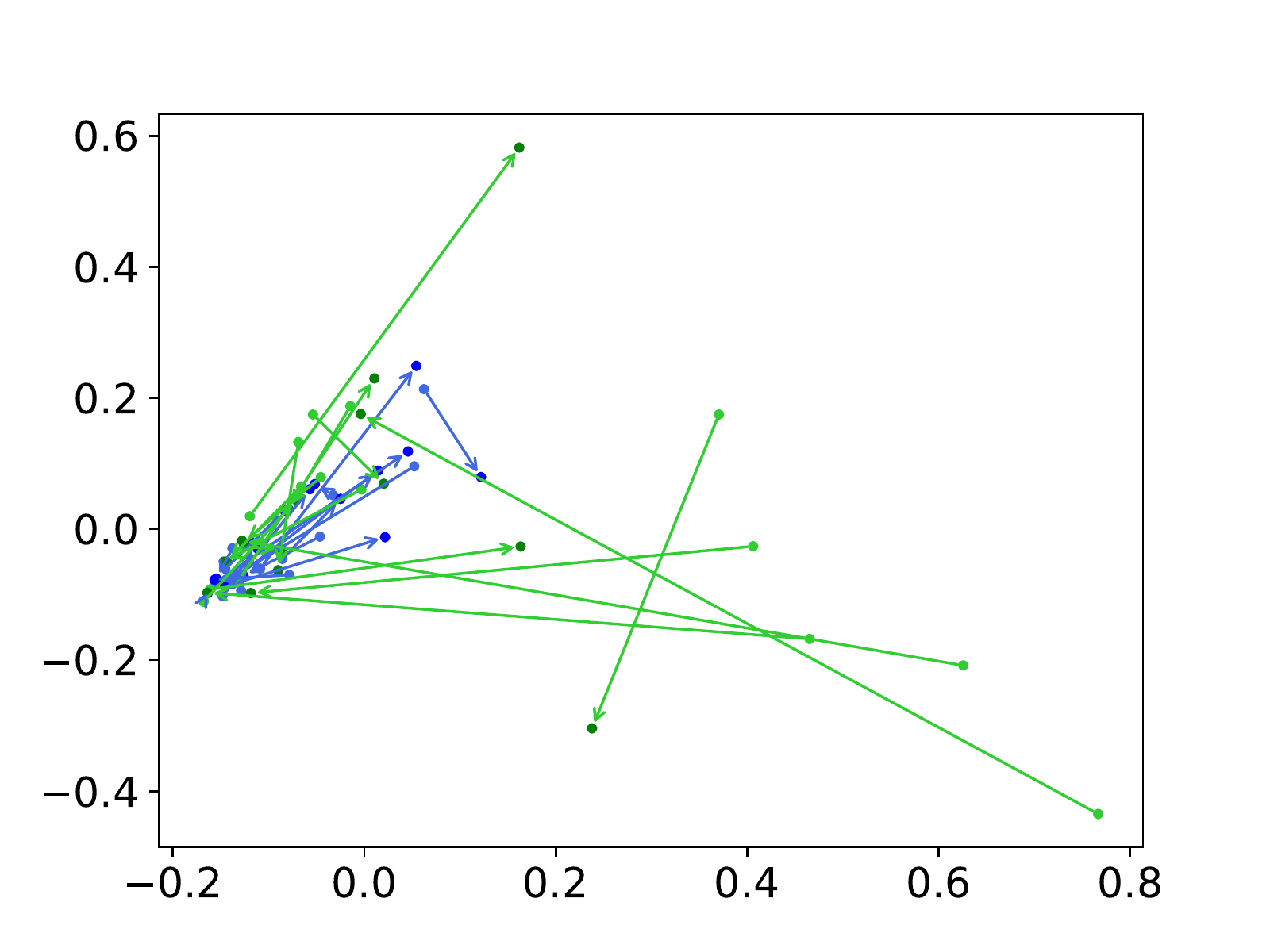}   
    \vspace*{-0.2cm}
  \end{minipage}
  \vspace*{-0.6cm}
  \caption{Three figures show the influence on the different levels under DDP, including the image space, the first and the fourth feature spaces of a ResNet18 respectively. In the first image, the first column is the original input and the remaining is the augmentation results of them. In the last two images, the blue and green arrow is the feature change of InD data and OOD data respectively under DDP after dimensionality reduction using PCA.}
  \label{fig_app_ddp_level}
  \vspace*{-0.2cm}
\end{figure}

In Figure \ref{fig_app_ddp_level}, we show the augmentation results of different inputs using DDP and the influence of DDP on the different feature spaces of a pre-trained ResNet18. The augmentation results have a more obvious semantic change for OOD input. For example, a rickshaw becomes a horse in Fig \ref{fig_app_ddp_level}.1. On the other hand, the semantic information of the augmentation results of the InD data is relatively unchanged. Correspondingly, We can find that the change in low-level features (the first feature space, in Fig \ref{fig_app_ddp_level}.2) is similar but the change in high-level semantic features (the fourth feature space, in Fig \ref{fig_app_ddp_level}.3) becomes small when the input is InD and relatively big when the input is OOD. 

Therefore, the low-level features concentrate on the pixel-level changes and do not differ much for the InD and OOD inputs. When we filter out the irrelevant information and focus on the semantic level changes, the change in the InD data is relatively small. This is because the interpolation operator can keep the InD input in the InD area. The effect of keeping the input in the InD area can only be observed in the high-level feature spaces. This explains why we cannot solve the OOD detection task with a single diffusion model and shows the information extraction ability of discriminator models.

\subsubsection{Different type of input}

\begin{figure}
  \centering
  \includegraphics[width=0.82\linewidth, trim=0 135 0 140]{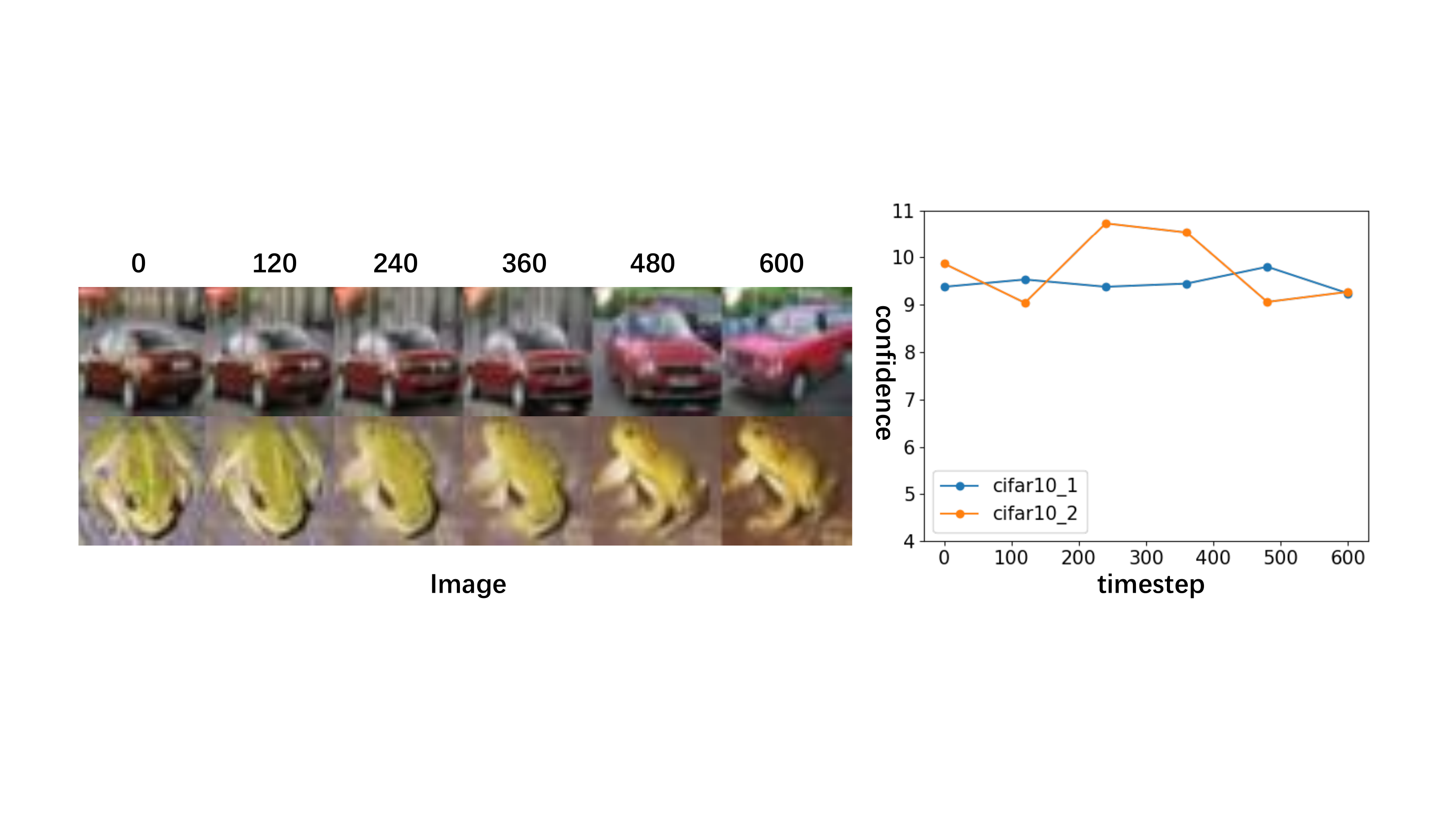}
  \includegraphics[width=0.82\linewidth, trim=0 135 0 140]{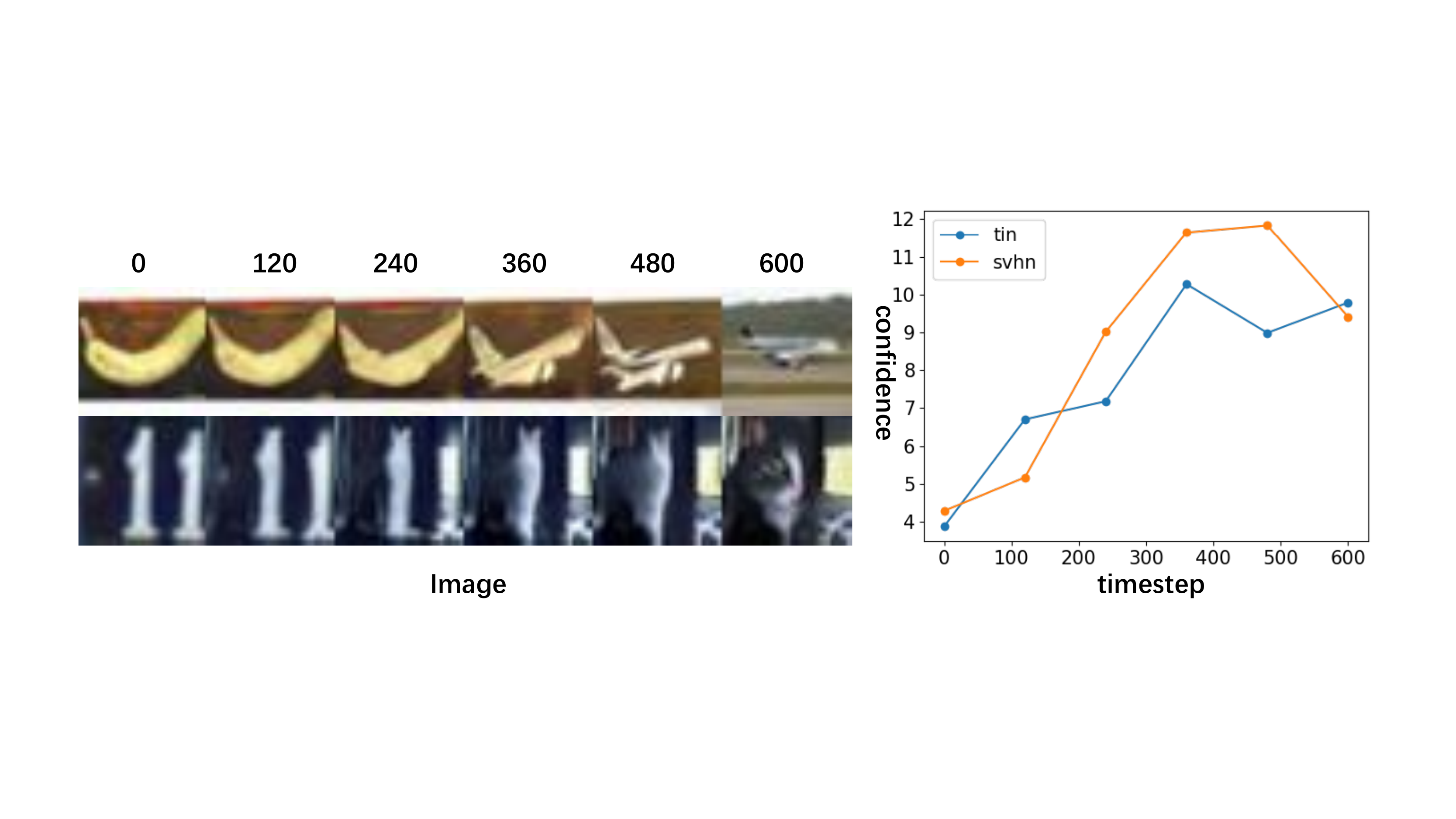}
  \includegraphics[width=0.82\linewidth, trim=0 160 0 140]{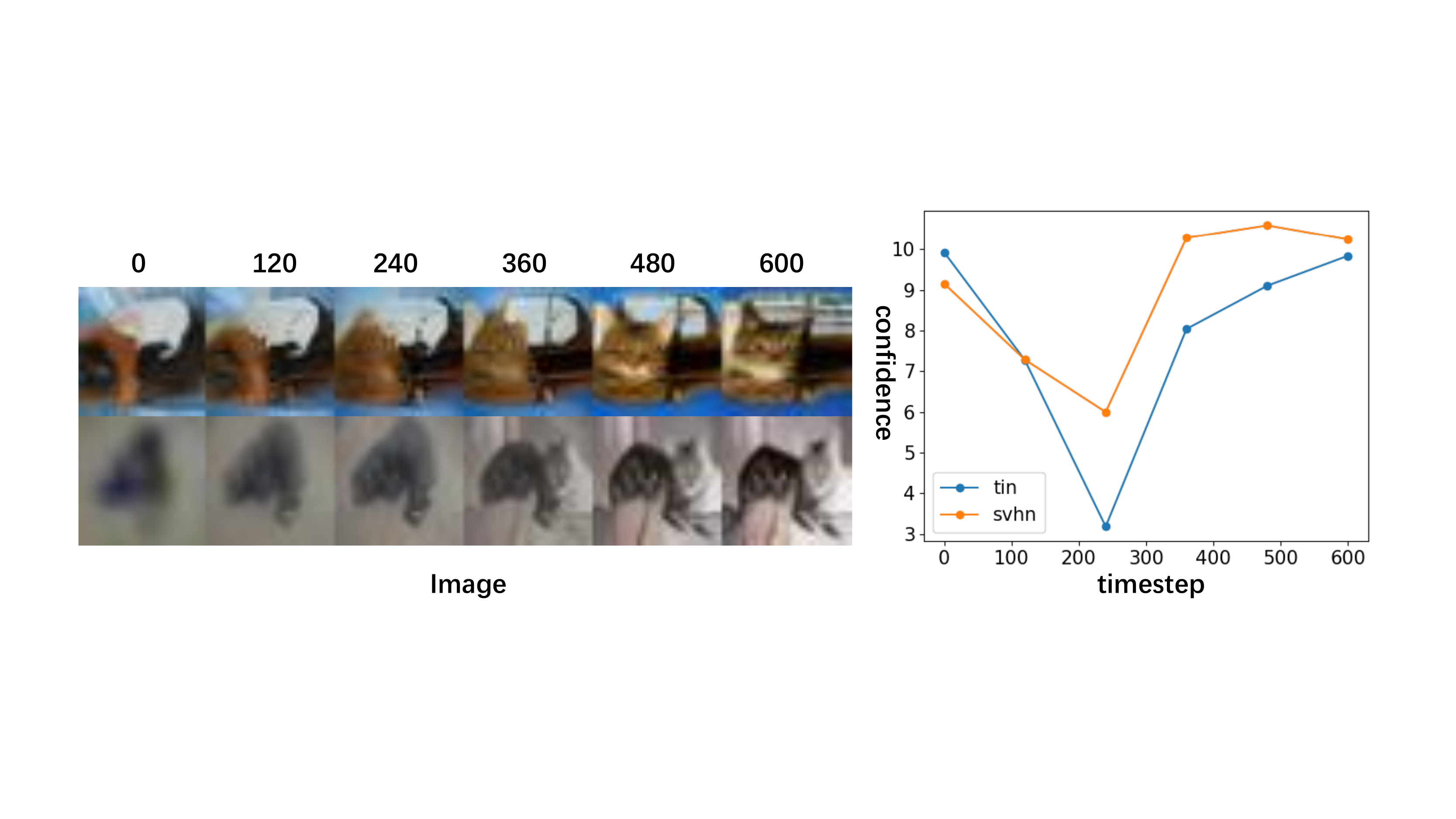}
  \caption{Changes in images under DDP and the corresponding confidence results.}
  \label{fig_app_input}
\end{figure}

In Figure \ref{fig_app_input}, we use real cases to show the relationship between different inputs and the maximum value of the logit (the confidence). For the first group, these InD data change in the InD area. Although the images change obviously in the pixel level under DDP, the confidence results remain largely stable. For the second group, these OOD data contain no class-related features and DDP pull them to the high-density area of the training data. Therefore, the confidence results increase generally. For the third group, these OOD data contain sparse class-related features, but these features are captured by the discriminator by coincidence. These are two examples of the overconfidence problem. However, these sparse class-related features cannot be maintained under DDP. There is a significant drop in the confidence results. All these results support the analysis in Section \ref{sec_main_method}.

\subsection{Experiment setting}
\label{app_exp_detail}

For CIFAR10/100, we employ a pre-trained ResNet18 \citep{he2016deep} and the diffusion model from DDPM. To obtain a conditional diffusion model, we add an additional condition embedding and concatenate it with the timestep embedding. For ImageNet, we use a pre-trained ResNet50 and a pre-trained latent diffusion model \citep{rombach2022high}. The main hyperparameters for our diffusion-based OOD detection:

\begin{table}[h]
  \centering
  \begin{tabular}{@{}llcc@{}}
  \toprule
  \multirow{2}{*}{Model} & \multirow{2}{*}{Hyperparam} & \multicolumn{2}{c}{Value} \\
   &  & CIFAR & ImageNet \\ \midrule
  \multirow{3}{*}{diffusion} & timestep & 300 & 600 \\
   & guidance scale & 3.0 & 6.0 \\ 
   & repeat size & 4 & 4 \\ \midrule
  \multirow{2}{*}{ResNet} & feature space & logit & feature5 \\
   & threshold & n/a & 0.3 \\ \bottomrule
  \end{tabular}
\end{table}

\subsection{FPR@95 result}
\label{app_fpr}

In Table \ref{tb_ood_f} and \ref{tb_imagenet_f}, we provide the FPR@95 results, which are highly consistent with the AUROC results in the main paper. We observe that the advantages in FPR@95 are more pronounced than in AUROC. The mean FPR@95 on CIFAR10 decreased from the previous SOTA value of 38.32 to 27.33, and the mean FPR@95 on CIFAR100 decreased from 75.12 to 59.08. This is because the FPR@95 is more sensitive to the OOD data with high confidence. The OOD data with high confidence are more likely to cause overconfidence problems. Our method can effectively reduce the confidence of OOD data, so the advantages are more obvious in FPR@95.

\begin{table}[h]
  \vspace*{-0.2cm}
  \centering
  \small
  \caption{The FPR@95 results of different methods on CIFAR10 and CIFAR100. The higher results are better and the bold results are the best in each case.}
  \vspace*{0.2cm}
  \label{tb_ood_f}
  \begin{tabular}{@{}l|ccccc|ccccc@{}}
    \toprule
    InD & \multicolumn{5}{c}{cifar10} & \multicolumn{5}{|c}{cifar100} \\
    OOD & cifar100 & tin & svhn & texture & place & cifar10 & tin & svhn & texture & place \\ \midrule
    ODIN & 58.72 & 55.52 & 67.46 & 50.53 & 50.07 & 83.1 & 77.67 & 89.72 & 78.37 & 81.2 \\
    EBO & 51.31 & 44.89 & 44.69 & 48.01 & 41.74 & 81.47 & 76.06 & 82.75 & 82.43 & 80.66 \\
    ReAct & 53.34 & 46.73 & 48.83 & 49.66 & 43.84 & 86 & 78.37 & 80.99 & 75.23 & 82.98 \\
    MLS & 52.04 & 45.37 & 44.55 & 48.56 & 42.72 & 81.27 & 75.25 & 82.39 & 82.46 & 80.22 \\
    VIM & 56.11 & 52.19 & \textbf{14.29} & \textbf{21.03} & 47.99 & 87.75 & 82.46 & 82.58 & \textbf{56.1} & 83.83 \\
    KNN & 52.17 & 46.36 & 33.43 & 45.6 & 43.4 & 82.27 & 73.82 & 74.37 & 66.4 & 78.74 \\ \midrule
    Diff(ours) & \textbf{35.95} & \textbf{28.52} & 19.6 & 23.75 & \textbf{28.82} & \textbf{71.53} & \textbf{52.38} & \textbf{49.78} & 59.55 & \textbf{62.18} \\ \bottomrule
    \end{tabular}
  \vspace*{-0.4cm}
\end{table}

\begin{table}[h]
  \vspace*{0.1cm}
  \centering
  \small
  \setlength\tabcolsep{5.75pt}
  \caption{The FPR@95 results of different methods on ImageNet and the mean and standard deviation of the results on CIFAR10, CIFAR100, and ImageNet. A higher mean value is better, while a lower standard variance is better. The best results are in bold.}
  \label{tb_imagenet_f}
  \vspace*{0.2cm}
  \begin{tabular}{@{}l|ccccc|cccccc@{}}
    \toprule
    InD & \multicolumn{5}{c}{imagenet} & \multicolumn{2}{|c}{cifar10} & \multicolumn{2}{c}{cifar100} & \multicolumn{2}{c}{imagenet} \\
    OOD/stat & species & inatur & texture & place & sun & mean & std & mean & std & mean & std \\ \midrule
    ODIN & 80.49 & 42.04 & \textbf{45.44} & 61.85 & 53.82 & 56.46 & 6.37 & 82.01 & 4.32 & 56.73 & 13.73 \\
    EBO & 82.61 & 53.73 & 52.43 & 66 & 58.8 & 46.13 & 3.26 & 80.67 & 2.42 & 62.71 & 11.03 \\
    ReAct & 68.53 & 19.36 & 45.6 & 33.47 & \textbf{24.02} & 48.48 & \textbf{3.15} & 80.71 & 3.71 & 38.20 & 17.62 \\
    MLS & 80.86 & 50.8 & 54.2 & 65.62 & 59.83 & 46.65 & 3.29 & 80.32 & 2.66 & 62.26 & \textbf{10.58} \\
    VIM & 85.72 & 73.01 & \textbf{13.05} & 85.61 & 83.7 & 38.32 & 17.20 & 78.54 & 11.38 & 68.22 & 27.98 \\
    KNN & 90.23 & 63.52 & 19.43 & 84.66 & 75.68 & 44.19 & 6.11 & 75.12 & 5.34 & 66.70 & 25.30 \\ \midrule
    Diff(ours) & \textbf{55.81} & \textbf{13.45} & 47.16 & \textbf{32.9} & \textbf{31.63} & \textbf{27.33} & 5.49 & \textbf{59.08} & 7.70 & \textbf{36.19} & 14.52 \\ \bottomrule
    \end{tabular}
  \vspace*{-0.2cm}
\end{table}

\subsection{Ablation study}
\label{app_ablation}

\begin{figure}[h]
  \centering
  \includegraphics[width=0.4\linewidth]{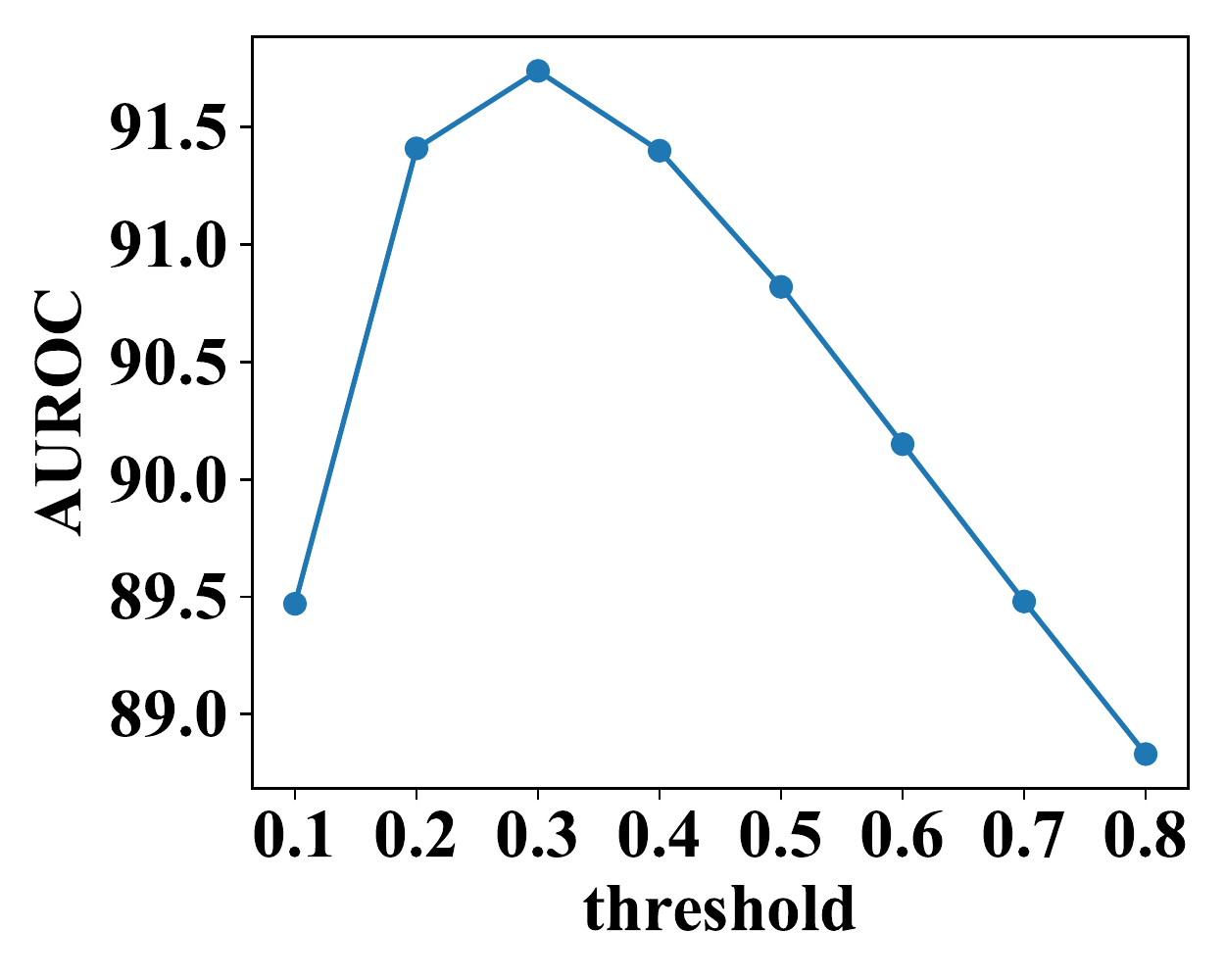}
  \vspace*{-0.2cm}
  \caption{the AUROC results obtained under different thresholds on ImageNet.}
  \label{fig_ab_thr}
\end{figure}

An interesting additional ablation study involves the threshold of features, which has been shown to be useful for the large ImageNet dataset in previous work ReAct. In Figure \ref{fig_ab_thr}, we find that 0.3 is the optimal choice for diffusion-based OOD detection, which differs from ReAct, where the optimal choice is around 1.0.

\newpage
\subsection{Project License}

Here, we present the GitHub repository addresses and the project licenses for the main open-source projects used in this paper.

\begin{table}[h]
  \centering
  \begin{tabular}{@{}lll@{}}
  \toprule
  name & GitHub & license \\ \midrule
  DDPM & \url{https://github.com/hojonathanho/diffusion} & n/a \\
  DDIM & \url{https://github.com/ermongroup/ddim} & MIT license \\
  Latent Diffusion & \url{https://github.com/CompVis/latent-diffusion} & MIT license \\
  OpenOOD & \url{https://github.com/Jingkang50/OpenOOD} & MIT license \\ \bottomrule
  \end{tabular}
\end{table}

\end{document}